\documentclass{egpubl}
 
\JournalSubmission    
\usepackage[T1]{fontenc}
\usepackage{dfadobe}

\usepackage{cite}  
\BibtexOrBiblatex

\electronicVersion
\PrintedOrElectronic
\ifpdf \usepackage[pdftex]{graphicx} \pdfcompresslevel=9
\else \usepackage[dvips]{graphicx} \fi

\usepackage{egweblnk} 

\title[3D Shape Generation: A Survey]%
      {3D Shape Generation: A Survey}

\author[Nicolas Caytuiro, Ivan Sipiran]{Nicolas Caytuiro, Ivan Sipiran}

\usepackage{booktabs}
\usepackage{array}
\usepackage{graphicx} 
\usepackage{makecell}

\newcolumntype{P}[1]{>{\raggedright\arraybackslash}p{#1}}

\newcommand{\tablaModelosGenerativos}{%
\begin{table*}[htp]
\centering
\caption{Representative 3D shape generation methods. The methods are first categorized according to the underlying generative paradigm and their associated generation space, which defines the 3D representation used for synthesis. We also report the datasets and metrics used for training and evaluation for each method. Please refer to Section~\ref{sec:metrics} for the definitions of the abbreviations used in the metrics column.}
\label{tab:models_summary}
\renewcommand{\arraystretch}{1.2}
\tiny
\resizebox{\textwidth}{!}{%
\begin{tabular}{c c c c c c}
\toprule
\textbf{Method} & \textbf{Generative Model} & \textbf{Generation Space} &
\textbf{Representation} & \textbf{Datasets} & \textbf{Metrics} \\
\midrule
3DGAN~\cite{wu_LearningProbabilisticLatent_2016} & GAN & Latent 1D 
& Voxel & ShapeNet, ModelNet, IKEA & Average precision \\

l-GAN/r-GAN~\cite{achlioptas_learning_2018} & GAN & Latent 1D 
& Point Cloud & ShapeNet & JSD, MMD, Cov, CD, EMD \\

GraphCNN-GAN~\cite{valsesia_LearningLocalizedGenerative_2018} & GAN & Latent 1D 
& Point Cloud & ShapeNet & JSD, MMD, Cov, CD, EMD \\

tree-GAN~\cite{shu_treeGAN3DPointCloud_2019} & GAN & Latent 1D 
& Point Cloud & ShapeNet & JSD, MMD, Cov, CD, EMD \\

SurfGen~\cite{luo_SurfGenAdversarial3D_2021} & GAN & Latent 1D 
& SDF & ShapeNet & JSD, MMD, Cov, CD, EMD \\

\midrule

SDM-Net~\cite{gao_SDMNETDeepGenerative_2019} & VAE & Latent 1D 
& Mesh & ShapeNet, ModelNet & JSD, MMD, Cov, CD, EMD \\

TM-Net~\cite{gao_TMNETDeepGenerative_2021} & VAE & Latent 1D 
& Mesh & ShapeNet, ModelNet & Image similarity measures \\ 

SetVAE~\cite{kim_SetVAELearningHierarchical_2021} & VAE & Latent 1D 
& Point Cloud & ShapeNet, Set-MNIST, Set-MultiMNIST & MMD, Cov, 1-NNA, CD, EMD \\

Michelangelo~\cite{zhao_MichelangeloConditional3D_2023} & VAE & Latent 1D 
& Mesh & ShapeNet & IoU, Other proposed by the authors \\

TAR3D~\cite{zhang_TAR3DCreatingHighQuality_2025} & VAE & Latent 3D 
& Triplane & ShapeNet, Objaverse, GSO & PSNR, LPIPS, SSIM, CLIP \\

COD-VAE~\cite{cho_Representing3DShapes_2025} & VAE & Latent 1D 
& Triplane & ShapeNet, Objaverse & IoU, CD, F-Score, FPD, MMD, Cov, 1-NNA \\

\midrule

PointFlow~\cite{yang_pointflow_2019} & Normalizing Flow & Latent 1D 
& Point Cloud & ShapeNet, ModelNet & JSD, MMD, Cov, 1-NNA, CD, EMD \\

DPF Networks~\cite{klokov_DiscretePointFlow_2020} & Normalizing Flow & Latent 1D 
& Point Cloud & ShapeNet & JSD, MMD, Cov, 1-NNA, CD, EMD \\

SoftFlow~\cite{kim_SoftFlowProbabilisticFramework_2020} & Normalizing Flow & Latent 1D 
& Point Cloud & ShapeNet & 1-NNA, CD, EMD \\

EAGLE~\cite{wang_EAGLEContextualPoint_2025} & Normalizing Flow & Latent 1D 
& Point Cloud & ShapeNet & JSD, MMD, Cov, 1-NNA, CD, EMD \\

\midrule

PointGrow~\cite{sun_PointGrowAutoregressivelyLearned_2020} & Autoregressive & Latent 3D 
& Point Cloud & ShapeNet, ModelNet, PASCAL3D+ & PointNet Distance (PND - Derived from FID) \\

PolyGen~\cite{nash_PolyGenAutoregressiveGenerative_2020} & Autoregressive & Polygon 
& Mesh & ShapeNet & Log-likelihood, CD \\

AutoSDF~\cite{mittal_AutoSDFShapePriors_2022} & Autoregressive & Latent 1D 
& SDF & ShapeNet, Pix3D & IoU, CD, F-Score \\

PointARU~\cite{meng_PointARU3DPoint_2025} & Autoregressive & Latent 3D 
& Point Cloud & ShapeNet & 1-NNA, CD, EMD \\

ShapeGPT~\cite{yin_ShapeGPT3DShape_2025} & Autoregressive & Latent 1D 
& Mesh & ShapeNet & IoU, CD, F-Score, P-FID, ULIP, CLIP \\

\midrule

PVD~\cite{zhou_3DShapeGeneration_2021} & Diffusion & Latent 3D 
& Point-Voxel & ShapeNet, PartNet & 1-NNA, CD, EMD \\

LION~\cite{zeng_LIONLatentPoint_2022} & Diffusion & Latent 3D 
& Point Cloud & ShapeNet & 1-NNA, Cov, MMD, CD, EMD \\

SALAD~\cite{koo_SALADPartLevelLatent_2023} & Diffusion & Latent 1D 
& Mesh & ShapeNet & 1-NNA, Cov, MMD, CD, EMD \\

SLIDE~\cite{lyu_SLIDEControllableMeshGeneration_2023} & Diffusion & Latent 3D 
& Point Cloud, Mesh & ShapeNet & 1-NNA, MMD, Cov, CD, EMD \\

DiffTF~\cite{cao_LargeVocabulary3DDiffusion_2023} & Diffusion & Latent 3D 
& Triplane & ShapeNet, OmniObject & FID, KID, Cov, MMD, CD \\

Sin3DM~\cite{wu_Sin3DMLearningDiffusion_2023} & Diffusion & Latent 3D 
& Triplane & Single 3D Textured Shape & G-Qual, SSFID, G-Div, IoU \\

XCube~\cite{ren_XCubeLargeScale3D_2024} & Diffusion & Latent Diffusion 
& Voxel & ShapeNet, Objaverse & 1-NNA, CD, EMD \\

TIGER~\cite{ren_TIGERTimeVaryingDenoising_2024} & Diffusion & Latent 3D 
& Point Cloud & ShapeNet & 1-NNA, CD, EMD \\

GPLD3D~\cite{dong_GPLD3DLatentDiffusion_2024} & Diffusion & Latent Diffusion 
& Mesh & ShapeNet & FPD, KPD \\

Direct3D~\cite{wu_Direct3DScalableImageto3D_2024} & Diffusion & Latent Diffusion 
& Mesh & Objaverse, GSO & CD, Volume IoU, F-Score \\

TetraDiffusion~\cite{kalischek_TetraDiffusionTetrahedralDiffusion_2025} & Diffusion & Tetrahedral 
& SDF & ShapeNet & 1-NNA, MMD, CD, EMD, FID, CLIP, KID \\

ShapeShifter~\cite{maruani_ShapeShifter3DVariations_2025a} & Diffusion & Latent 3D 
& Voxel, Points & Single 3D Textured Shape & SSFID, IoU \\

CoPart~\cite{dong_OneMoreContextual_2025} & Diffusion & Latent 3D, 2D 
& Mesh & PartVerse & CLIP, ULIP \\

Hunyuan3D 2.1~\cite{hunyuan3d_Hunyuan3D21Images_2025} & Diffusion & Latent Diffusion 
& Mesh & ShapeNet, ModelNet40, Thingi10K, Objaverse & ULIP, Uni3D \\

VolGen~\cite{tang_VolGenVolumetricLatent_2025} & Diffusion & Latent 3D 
& Mesh & Objaverse & CD, IoU, ULIP, Uni3D \\

Hi3DGen~\cite{ye_Hi3DGenHighfidelity3D_2025} & Diffusion & Latent Diffusion 
& Mesh, Normal map & DetailVerse & Image-to-normal estimation metrics \\

\bottomrule
\end{tabular}
}
\end{table*}
}

\newcommand{\tableTrainingTimesComp}{%
\begin{table*}[htp]
\centering
\caption{Representative 3D Generative Models and their reported training/inference times. For dataset references associated with each model, please refer to table \ref{tab:models_summary}. For some methods, training and inference times are not explicitly reported; in those cases, we provide estimated averages derived from the model architecture, dataset scale, and GPU details.}
\label{tab:trainingTimeComp}
\renewcommand{\arraystretch}{1.2}
\tiny
\resizebox{\textwidth}{!}{%
\begin{tabular}{c c c c c}
\toprule
\textbf{Method} & \textbf{GPU Details} & \textbf{$\approx$ Training Time (GPU hours)} &
\textbf{$\approx$ Inference Time (s)} \\
\midrule

GraphCNN-GAN~\cite{valsesia_LearningLocalizedGenerative_2018} & NVIDIA Quadro P6000 & 120 & 0.1 \\

SDM-Net~\cite{gao_SDMNETDeepGenerative_2019} & 1 $\times$ GTX 1080 Ti GPU & 120 & 0.1 \\

TM-Net~\cite{gao_TMNETDeepGenerative_2021} & 12 $\times$  RTX 2080Ti GPUs & 120 & 0.80 \\ 

Michelangelo~\cite{zhao_MichelangeloConditional3D_2023} & 8 $\times$ NVIDIA Tesla V100 GPUs & 120 & 10 \\

TAR3D~\cite{zhang_TAR3DCreatingHighQuality_2025} & 8 $\times$ NVIDIA A100 GPUs & 120 & 17.7 \\

COD-VAE~\cite{cho_Representing3DShapes_2025} & 8 $\times$ RTX 4090 GPUs, 4 $\times$ A6000 GPUs & 408 & 0.1 \\

PointARU~\cite{meng_PointARU3DPoint_2025} & 8 $\times$ NVIDIA Tesla V100 GPUs & 125 & 3.21 \\

ShapeGPT~\cite{yin_ShapeGPT3DShape_2025} & 4 $\times$ NVIDIA A100 GPUs & 600 & 0.6 \\

PVD~\cite{zhou_3DShapeGeneration_2021} & 1 $\times$ NVIDIA Titan RTX & 142 & 8.46 \\

LION~\cite{zeng_LIONLatentPoint_2022} & In-house GPU cluster of NVIDIA V100 GPUs & 550 & 27.12 \\

SLIDE~\cite{lyu_SLIDEControllableMeshGeneration_2023} & In-house GPU cluster of NVIDIA V100 GPUs & 360 & 0.2 \\

TIGER~\cite{ren_TIGERTimeVaryingDenoising_2024} & In-house GPU cluster of NVIDIA V100 GPUs & 164 & 9.73 \\

CoPart~\cite{dong_OneMoreContextual_2025} & 4 $\times$ NVIDIA V100 GPUs & 120 & 65 \\

Hunyuan3D 2.1~\cite{hunyuan3d_Hunyuan3D21Images_2025} & In-house GPU cluster of NVIDIA V100 GPUs & 180 & 60 \\

VolGen~\cite{tang_VolGenVolumetricLatent_2025} & 16 $\times$ NVIDIA A800 GPUs & 504 & 10 \\

Hi3DGen~\cite{ye_Hi3DGenHighfidelity3D_2025} & 8 $\times$ NVIDIA A800 GPUs & 500 & 10 \\

\midrule

SSG~\cite{wu_LearningGenerate3D_2022a} & 1 $\times$ NVIDIA 3090 GPU, 1 $\times$ NVIDIA 2070 GPU & 4 & 0.1 \\

Sin3DGen~\cite{li_PatchBased3DNatural_2023} & 1 $\times$ NVIDIA V100 GPU & 0.25 & 15.8 \\

Sin3DM~\cite{wu_Sin3DMLearningDiffusion_2023} & 1 $\times$ NVIDIA RTX A6000 GPU & 3 & 15.8 \\

ShapeShifter~\cite{maruani_ShapeShifter3DVariations_2025a} & 1 $\times$ NVIDIA GeForce RTX 3080 GPU & 0.2 & 10.7 \\

\bottomrule
\end{tabular}
}
\end{table*}
}

\usepackage{amsmath}
\usepackage{amsfonts}
\usepackage{amssymb}

\begin{document}


\teaser{
 \includegraphics[width=1\linewidth]{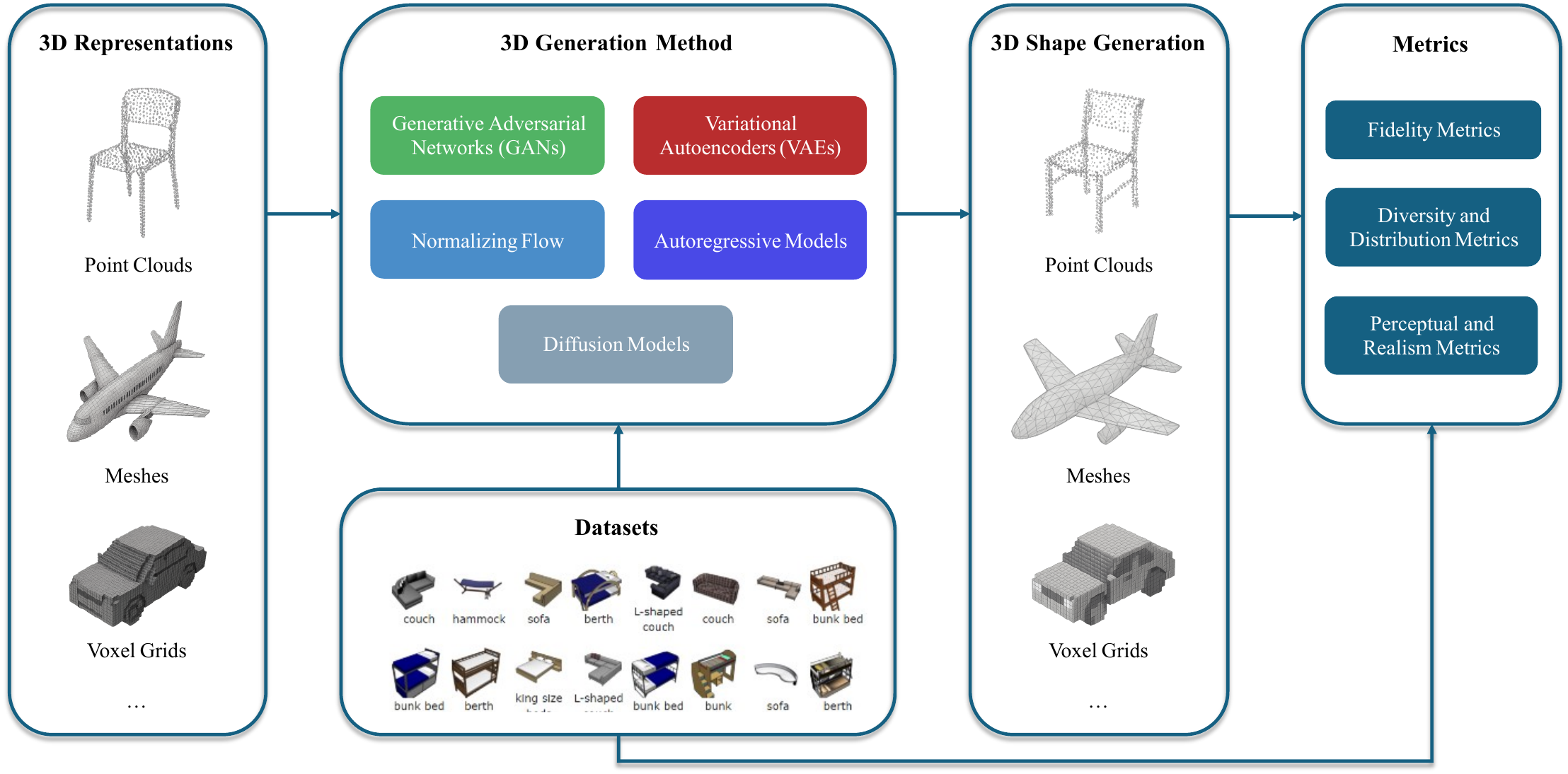}
 \centering
  \caption{Overview of this survey, covering 3D representations, generation methods, datasets, and evaluation metrics. We first introduce the 3D representations that serve as the foundation for 3D shape generation. Next, we provide a comprehensive overview of the rapidly expanding body of literature on generation methods, focusing specifically on feedforward paradigms. Finally, we briefly describe the most widely used datasets for 3D generation and the metrics employed to evaluate the quality of the generated shapes.}
\label{fig:overview_survey}
}

\maketitle
\begin{abstract}
   Recent advances in deep learning have significantly transformed the field of 3D shape generation, enabling the synthesis of complex, diverse, and semantically meaningful 3D objects. This survey provides a comprehensive overview of the current state-of-the-art in 3D shape generation, organizing the discussion around three core components: shape representations, generative modeling approaches, and evaluation protocols. We begin by categorizing 3D representations into explicit, implicit, and hybrid setups, highlighting their structural properties, advantages, and limitations. Next, we review a wide range of generation methods, focusing on feedforward architectures. We further summarize commonly used datasets and evaluation metrics that assess fidelity, diversity, and realism of generated shapes. Finally, we identify open challenges and outline future research directions that could drive progress in controllable, efficient, and high-quality 3D shape generation. This survey aims to serve as a valuable reference for researchers and practitioners seeking a structured and in-depth understanding of this rapidly evolving field.

\ccsdesc[500]{Computing methodologies~Computer vision}

\printccsdesc   
\end{abstract}  

\tablaModelosGenerativos

\section{Introduction}

Generative modeling of 3D shapes has gained significant attention due to its broad applicability across diverse domains, including medicine~\cite{guo_DeepLearningNetwork_2018, friedrich_PointCloudDiffusion_2023a, durrer_DenoisingDiffusionModels_2025}, robotics~\cite{ni_PointNetGraspingLearning_2020}, bioinformatics~\cite{hoogeboom_EquivariantDiffusionMolecule_2022, xu_GeoDiffGeometricDiffusion_2021, you_Latent3DGraph_2023}, computer-aided design~\cite{hertz_SPAGHETTIEditingImplicit_2022, li_EditVAEUnsupervisedPartsAware_2022, hu_CNSEdit3DShape_2024, lee_PASTAPartAwareSketchto3D_2025}, cultural heritage~\cite{hermoza_3DReconstructionIncomplete_2018, xu_CPDCMFNetConditionalPoint_2024}, and more. To support such downstream tasks, a successful 3D generative model should be both \emph{faithful} and \emph{probabilistic}~\cite{zhou_3DShapeGeneration_2021}. A faithful model generates geometrically plausible and visually realistic shapes, while also respecting partial observations (e.g., depth maps or occlusions) when conditioned. On the other hand, a probabilistic model must capture the inherent ambiguity and multimodality of 3D shape generation, allowing diverse outputs from incomplete or ambiguous inputs. 

Given the data-intensive nature of deep learning, the construction and benchmarking of large-scale 3D shape datasets have become essential for training and evaluating generative models~\cite{chang_ShapeNetInformationRich3D_2015, zhirongwu_3DShapeNetsDeep_2015}.

This survey provides a comprehensive overview of deep learning techniques for 3D shape generation (see Table \ref{tab:models_summary}). Compared to traditional 3D acquisition and modeling techniques--which emphasize precision and geometric accuracy through handcrafted features and algorithmic pipelines--deep learning-based 3D shape generation offers the advantage of learning complex, high-dimensional latent spaces that capture both structural and semantic properties of 3D objects. This enables not only the synthesis of novel and diverse shapes, but also intuitive manipulation in the latent space, such as interpolation, extrapolation, and conditional editing.

However, learning to generate 3D shapes introduces unique challenges. Thus, the complexity and irregularity of 3D shapes and the different requirements of practical applications have resulted in the lack of a unified representation of 3D shapes~\cite{xiao_SurveyDeepGeometry_2020}. Instead, various 3D representations are employed, each with its own structural properties, advantages, and limitations. These factors directly impact model design, data processing, and evaluation strategies.

We organize the discussion around three core components of the 3D shape generation pipeline: representations, generation methods, and evaluation protocols.
Figure \ref{fig:overview_survey} presents an overview of this survey. We first discuss the scope and related works in Section \ref{sec:scope_survey}. Section \ref{section_3d_shape_representations} introduces major categories of 3D shape representations: explicit, implicit, and hybrid. Section \ref{3d_generation_methods} explores the most relevant 3D shape generation methods, with a focus on the feedforward generative models. Sections \ref{sec:datasets} and \ref{sec:metrics} present widely used datasets and metrics for assessing the fidelity and diversity of generated shapes, respectively. Finally, in Section \ref{sec:discussion}, we discuss the open challenges and future directions. 

We hope this survey offers a systematic summary of 3D shape generation, serving as a valuable reference for both researchers and newcomers in this field.

\section{Scope of this Survey}\label{sec:scope_survey}

In this survey, we focus on techniques for 3D shape generation, along with the associated datasets and evaluation metrics. We begin by describing 3D representations, and then discuss how these representations integrate with various generative models. Subsequently, we provide a comprehensive overview of the main approaches within the feedforward generation paradigm. Finally, we review the datasets and metrics commonly used to evaluate 3D shape generation models.

\begin{figure*}[tbp]
\centering
\includegraphics[width=1\textwidth]{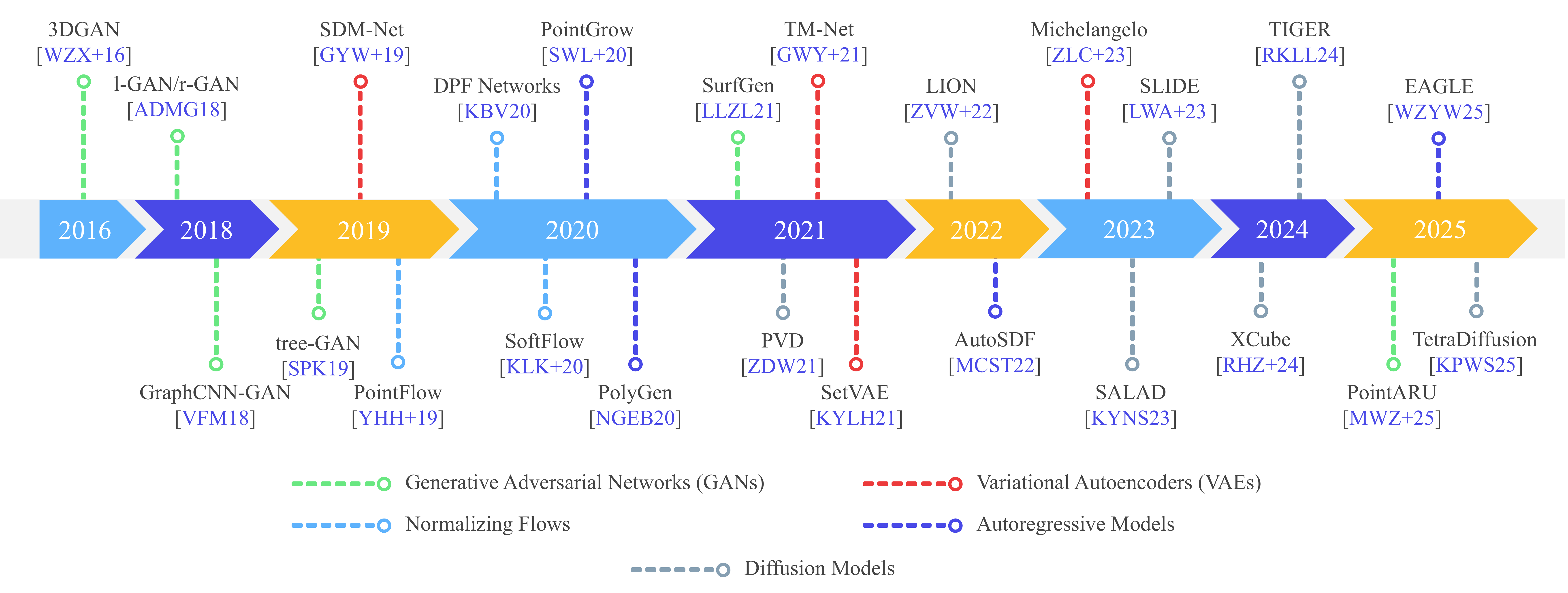}
\caption{Timeline of representative 3D shape generation models.}
\label{fig:timeline_models}
\end{figure*}

This survey aims to systematically summarize and categorize methods for 3D shape generation, together with the relevant representations, datasets, and metrics. The surveyed papers are primarily published in top-tier computer vision and computer graphics conferences and journals, as well as selected preprints published on \emph{arXiv} in 2025. While it is challenging to exhaustively cover all methods related to 3D shape generation, we strive to include the most significant branches of the field. Rather than providing exhaustive explanations for each branch, we highlight representative works (as shown in Figure \ref{fig:timeline_models}) to illustrate their main contributions and limitations. Further details can be found in the related work of these cited papers.

\subsection{Related Survey}

Our primary focus is on exploring techniques that generate 3D shapes. Therefore, this review does not encompass research on generation methods in the 2D domain. Several recent surveys have addressed specific aspects of 3D generation, including Deep Learning for 3D point cloud~\cite{guo_DeepLearning3D_2021}, 3D-aware image synthesis~\cite{xia_SurveyDeepGenerative_2023}, 3D generative models~\cite{shi_DeepGenerativeModels_2023}, Text-to-3D~\cite{li_GenerativeAIMeets_2024, lee_Textto3DShapeGeneration_2024}, 3D generation methods based on scene representation~\cite{li_Advances3DGeneration_2024}, implicit representations-based 3D shape generation~\cite{sun_RecentAdvancesImplicit_2024}, and diffusion models for 3D generation~\cite{wang_DiffusionModels3D_2025, kumar_DiffusionModelsGenerative_2025}. In contrast, this survey provides an in-depth and comprehensive review of 3D shape generation methods and their related datasets and metrics (as shown in Table \ref{tab:models_summary}). 

\section{3D Shape Representations}\label{section_3d_shape_representations}

3D shape representations are central to 3D computer vision and graphics~\cite{peng_ShapePointsDifferentiable_2021}. Shape representations can generally be categorized as either \emph{explicit} or \emph{implicit}. Explicit shape representations and learning algorithms depending on such representations directly parameterize the surface of the geometry, either as a point cloud~\cite{fan_PointSetGeneration_2017, ma_PowerPointsModeling_2021, qi_PointNetDeepLearning_2017, yang_pointflow_2019, wu_FastPointCloud_2023, meng_PointARU3DPoint_2025}, meshes~\cite{wang_Pixel2MeshGenerating3D_2018, jiang_SphericalCNNsUnstructured_2019, gupta_NeuralMeshFlow_2020}, or surface patches~\cite{groueix_AtlasNetPapierMacheApproach_2018, williams_DeepGeometricPrior_2019}.

While explicit representations are often computationally efficient and require few parameters, they tend to suffer from several limitations: discretization artifacts, difficulties in representing watertight surfaces, and constraints imposed by fixed or pre-defined topologies (meshes). In contrast, implicit representations define shapes as level sets of continuous functions over 3D space. These functions can be represented using discretized voxel grids~\cite{dai_ScanCompleteLargeScaleScene_2018, wu_LearningProbabilisticLatent_2016} or, more recently, parameterized using neural networks, commonly referred to as neural implicit functions~\cite{ye_GIFSNeuralImplicit_2022, burkov_MultiNeuS3DHead_2023}.

To bridge the gap between these two paradigms, a new class of methods known as \emph{hybrid} representations has emerged. These approaches aim to combine the structural efficiency of explicit representations with the flexibility and continuity of implicit ones, enabling more expressive and plausible 3D shape modeling.

\subsection{Explicit Representations}

\subsubsection{Point Clouds}

According to~\cite{park_DeepSDFLearningContinuous_2019}, a point cloud is a lightweight 3D representation that closely matches the raw data that many sensors (e.g., LiDARs, depth cameras) provide, and hence is a natural fit for applying 3D learning.

In this context,~\cite{qi_PointNetDeepLearning_2017} represents a point cloud as an unordered set of 3D points \(\{P_i \mid i = 1, \dots, n\}\), where each point $P_i \in \mathbb{R}^n$ is a feature vector that includes its 3D coordinates $(x,y,z)$, and optional attributes such as color, normal, or intensity values. In the basic case where only the spatial coordinates are used, $n = 3$. 

A primary limitation of point clouds is that they do not describe the topology and are not suitable for producing watertight surfaces~\cite{park_DeepSDFLearningContinuous_2019}. Another issue with point clouds as a representation is that they are unordered, which means that any permutation of a point set still describes the same shape. This complicates comparisons between two point sets, typically needed to define a loss function~\cite{achlioptas_learning_2018}.

\subsubsection{Meshes}

According to~\cite{botsch_PolygonMeshProcessing_2010, bronstein_GeometricDeepLearning_2017, wang_Pixel2MeshGenerating3D_2018}, a 3D mesh is a collection of vertices, faces, and edges that defines the shape of a 3D object; it can be represented by a graph structure (simplicial complex) \( \mathcal{M} = (\mathcal{V}, \mathcal{F}, \mathcal{E}) \), where 

\begin{equation*}
    \mathcal{V} = \{v_1,\dots,v_V\}
\end{equation*}

 is a set of vertices in the mesh, and 

 \begin{equation*}
      \mathcal{F} = \{f_1,\dots,f_F\}, \quad f_i \in \mathcal{V} \times \mathcal{V} \times \mathcal{V}
 \end{equation*}

 is a set of triangular faces connecting them. However, it is more efficient to represent the connectivity of a triangle mesh in terms of the edges of the respective graph~\cite{botsch_PolygonMeshProcessing_2010},

 \begin{equation*}
    \mathcal{E} = \{e_1,\dots,e_E\}, \quad e_i \in \mathcal{V} \times \mathcal{V}.
 \end{equation*}
 
In this sense, a mesh-based representation stores the surface information cheaply as a list of vertices and faces that respectively define the geometric and topological information. Early approaches to mesh generation focused on predicting the parameters of category-specific mesh models~\cite{zuffi_LionsTigersBears_2018, kolotouros_ConvolutionalMeshRegression_2019}. Although these methods produce manifold meshes, they are limited to object categories for which parameterized templates are available. Building upon this idea, works such as~\cite{groueix_AtlasNetPapierMacheApproach_2018, wang_Pixel2MeshGenerating3D_2018} extended mesh generation to a wider variety of object categories by incorporating topological priors, enabling more flexible and expressive 3D shape modeling beyond category-specific templates.

Compared to voxels, meshes are more compact and better suited to representing finer surface details. Compared to points, meshes are more controllable and exhibit better visual quality~\cite{gao_SDMNETDeepGenerative_2019}.

\subsubsection{Voxel Grids}

The direct extension of pixels in 2D images to 3D is the voxel representation, which has a regular grid structure well-suited for convolutional neural networks (CNNs)~\cite{girdhar_LearningPredictableGenerative_2016, wu_LearningProbabilisticLatent_2016, gao_SDMNETDeepGenerative_2019}.

However, voxel-based approaches suffer from severe scalability issues: both memory and computation grow cubically with the resolution of the output grid. As a result, training networks at resolutions higher than $64^3$ becomes impractical due to GPU memory limitations, often requiring strategies such as reducing batch size or generating volumetric output in smaller volumes~\cite{tatarchenko_OctreeGeneratingNetworks_2017}. Additionally, higher resolutions significantly increase training times, further limiting applicability.

Moreover, as highlighted by~\cite{dai_ScanCompleteLargeScaleScene_2018}, small voxel sizes capture local detail but lack spatial context; large voxel sizes provide large spatial context but lack local detail.

\subsection{Implicit Representations}

\subsubsection{Neural Radiance Fields (NeRF)}

\cite{mildenhall_NeRFRepresentingScenes_2022} introduce NeRF, a method for representing a 3D scene as an implicit function defined as 

\begin{equation}
F_\Theta : (\mathbf{x}, \mathbf{d}) \mapsto (\mathbf{c}, \sigma)
\end{equation}

where $\mathbf{x}$ is a 3D spatial coordinate, $\mathbf{d}$ is a 3D viewing direction, $\mathbf{c}$ is an RGB color, and $\sigma$ is a non-negative density value.

NeRFs model the appearance and geometry of 3D shapes by using volume rendering (ray matching) algorithms and positional encoding~\cite{mildenhall_NeRFRepresentingScenes_2022, metzer_LatentNeRFShapeGuidedGeneration_2023}. This technique can efficiently reconstruct geometry given only a few multiview posed images. To achieve higher geometry quality, NeuS~\cite{wang_NeuSLearningNeural_2021} and VolSDF~\cite{yariv_VolumeRenderingNeural_2021} extend radiance fields by using SDF instead of density as the geometry~\cite{sun_RecentAdvancesImplicit_2024}.

\subsubsection{Signed Distance Field (SDF)}

According to~\cite{jun_ShapEGeneratingConditional_2023}, Signed Distance Fields (SDFs) are a classic way to represent a 3D shape as a scalar field. In particular, an SDF $\mathbf{}{f}$ maps a coordinate $\mathbf{x}$ to a scalar $\mathbf{}{f}(\mathbf{x})=d$, such that $|d|$ is the distance of $\mathbf{x}$ to the nearest point on the surface of the shape. As a result of this definition, the level set $\mathbf{}{f}(\mathbf{x})=0$ defines the boundary of the shape, and sign($d$) determines normal orientations along the boundary. 

As a function, SDF is more flexible than the common explicit representations, such as point clouds or meshes, and inherently allows topology manipulations such as constructive solid geometry (CSG) operations. Moreover, SDF allows the use of a technique known as ``sphere tracing"~\cite{hart_SphereTracingGeometric_1996}, which can accelerate the rendering of path tracing~\cite{sun_RecentAdvancesImplicit_2024}.

Methods such as marching cubes~\cite{lorensen_MarchingCubesHigh_1987} or marching tetrahedra~\cite{doi_EfficientMethodTriangulating_1991} can be used to construct meshes from this level set.

\subsubsection{Neural Implicit Functions (NIFs)}

Neural implicit functions (NIFs) have recently gained significant attention as a powerful and flexible framework for representing 3D scenes using neural networks. Unlike traditional explicit representations such as meshes, NIFs are not constrained by fixed spatial resolution, allowing them to capture fine geometric details at arbitrary scales~\cite{burkov_MultiNeuS3DHead_2023}. 

A recent contribution in this domain is GIFS (General Implicit Function for 3D Shape), proposed by~\cite{ye_GIFSNeuralImplicit_2022}, which introduces a novel implicit representation along with an algorithm for extracting explicit surfaces. Instead of partitioning 3D shapes into predefined categories, GIFS models the pairwise relationships between 3D points. Specifically, it defines a binary flag that indicates whether two points lie on the same side of an object's surface. The method embeds 3D points into a latent space and employs neural networks to approximate these binary flags based on the point embeddings. In doing so, the approach is able to implicitly separate spatial regions, with the object shape emerging as the decision boundary defined by the binary classifications. 

\subsection{Hybrid Representations}

\subsubsection{Point-Voxel}

Given that voxel grids are memory-intensive, scaling cubically with resolution, they are unsuitable for high-resolution 3D shape generation. On the other hand, most point cloud processing networks are inherently permutation-invariant, which imposes a strong structural constraint on the model architecture~\cite{ibing_3DShapeGeneration_2021}.

To address these limitations,~\cite{ibing_3DShapeGeneration_2021} proposes the use of point-voxel representations for 3D shape generation. This hybrid representation had been previously explored in methods such as PointCNN~\cite{li_PointCNNConvolutionXTransformed_2018}, PV-RCNN~\cite{shi_PVRCNNPointVoxelFeature_2020}, and PVD~\cite{zhou_3DShapeGeneration_2021}. The core idea behind point-voxel is to voxelize the point clouds to enable efficient 3D convolutions while preserving point-level detail for finer geometric modeling.

\subsubsection{Tri-Plane}

In~\cite{chan_EfficientGeometryaware3D_2022}, the Tri-Plane representation was introduced as a hybrid explicit-implicit 3D representation that balances computational efficiency with representational expressiveness.

\begin{figure*}[h!]
\centering
\includegraphics[width=0.8\textwidth]{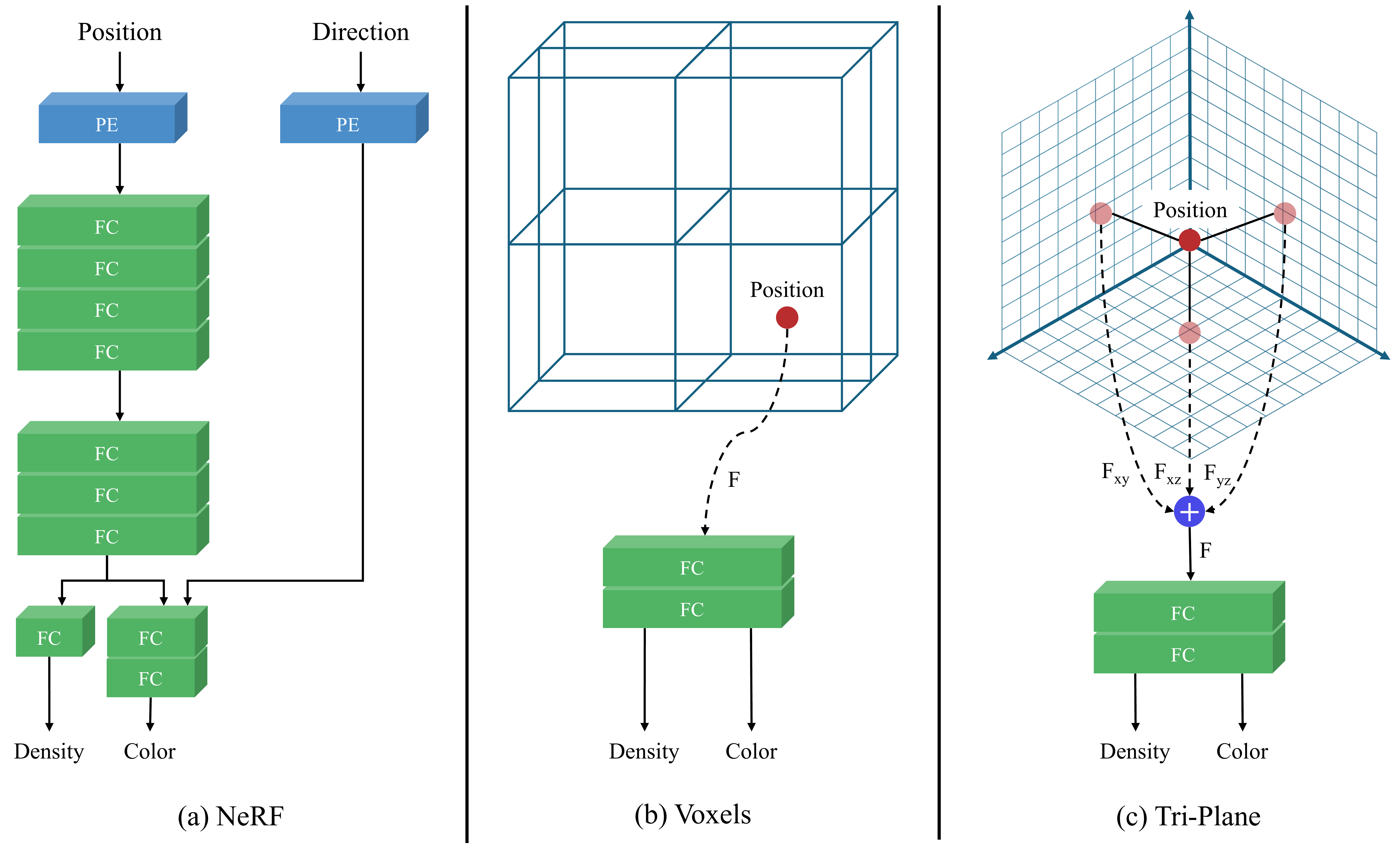}
\caption{Comparison of neural implicit, voxel-based, and tri-plane representations.
(a) Neural implicit representations rely on fully connected (FC) layers with positional encoding (PE). They provide high fidelity but suffer from slow query times.
(b) Explicit voxel grids, as well as hybrid variants with local decoders, allow for faster queries, yet their memory and computational costs grow rapidly with resolution.
(c) Tri-plane representations strike a balance by supporting efficient queries while scaling more gracefully with resolution.}
\label{fig:triplane_comparison}
\end{figure*}

In this formulation, features are explicitly arranged on three orthogonal, axis-aligned 2D planes--namely the $xy$, $yz$, and $xz$ planes, each with a resolution $N \times N \times C$, where $N$ denotes spatial resolution and $C$ the number of feature channels. As illustrated in Fig.~\ref{fig:triplane_comparison}, this structure contrasts with traditional 3D representations by encoding volumetric information through multiple planar projections.

To query a 3D point $x \in \mathbb{R}^3$, it is projected onto each of the three feature planes. The corresponding feature vectors $F_{xy}(\mathbf{x})$, $F_{xz}(\mathbf{x})$, and $F_{yz}(\mathbf{x})$ are retrieved using bilinear interpolation. These are then aggregated via element-wise summation, forming a unified feature vector.

\begin{equation}\label{eq:triplane}
\text{NF}(\mathbf{x}) = \text{MLP} \left( \mathbf{f}_{xy}(\mathbf{x}) + \mathbf{f}_{yz}(\mathbf{x}) + \mathbf{f}_{xz}(\mathbf{x}) \right). 
\end{equation}

A lightweight decoder, typically implemented as a small MLP, maps the aggregated features to radiance and density values, which are then rendered into RGB images using neural volume rendering techniques such as NeRFs.

In contrast to the original formulation,~\cite{shue_3DNeuralField_2023} confirmed that simple addition (as shown in Equation \ref{eq:triplane}) is an effective and efficient aggregation strategy for tri-plane features, avoiding the complexity of more elaborate fusion mechanisms.

The primary advantage of tri-planes representation lies in its efficiency--by keeping the decoder small and shifting the bulk of the expressive power into the explicit features. As demonstrated by~\cite{chan_EfficientGeometryaware3D_2022}, this approach reduces the computational cost of neural rendering compared to fully implicit MLP architectures without losing expressiveness.

\section{3D Generation Methods}\label{3d_generation_methods} 

According to~\cite{koo_SALADPartLevelLatent_2023}, the first 3D generative models are based on GANs, learning a distribution of latents that can be decoded into various 3D representations such as point clouds~\cite{achlioptas_learning_2018, valsesia_LearningLocalizedGenerative_2018, shu_treeGAN3DPointCloud_2019} and implicit representations~\cite{kleineberg_AdversarialGenerationContinuous_2020, hao_DualSDFSemanticShape_2020, chen_LearningImplicitFields_2019, ibing_3DShapeGeneration_2021, zheng_SDFStyleGANImplicitSDFBased_2022}. Building upon this foundation, the rapid development of generative models introduces new generation methods such as flow-based models~\cite{yang_pointflow_2019, klokov_DiscretePointFlow_2020}, which learn invertible mappings between complex and straightforward point distributions, enabling exact likelihood estimation and diverse point cloud synthesis; variational autoencoders (VAEs)~\cite{groueix_AtlasNetPapierMacheApproach_2018}, which encode 3D shapes into a probabilistic latent space and reconstruct them via learned decoders, allowing interpolation and latent sampling; autoregressive models~\cite{sun_PointGrowAutoregressivelyLearned_2020, mo_StructureNetHierarchicalGraph_2019}, which generate 3D geometry sequentially--either point-by-point or part-by-part--capturing structural dependencies; and, more recently, diffusion-based approaches~\cite{zeng_LIONLatentPoint_2022, nichol_PointESystemGenerating_2022, hunyuan3d_Hunyuan3D21Images_2025}, which progressively refine noisy inputs into clean 3D shapes through iterative denoising steps. Table \ref{tab:models_summary} shows well-known examples of 3D generation methods using generative models. Table \ref{tab:models_summary} also reports the metrics and datasets used to evaluate each generation method.

In this section, we explore the feedforward generation paradigm, proposed by~\cite{li_Advances3DGeneration_2024}, which enables the direct synthesis of 3D representations through generative models. 

\subsection{Feedforward Generation}

\begin{figure*}[tbp]
\centering
\includegraphics[width=1\textwidth]{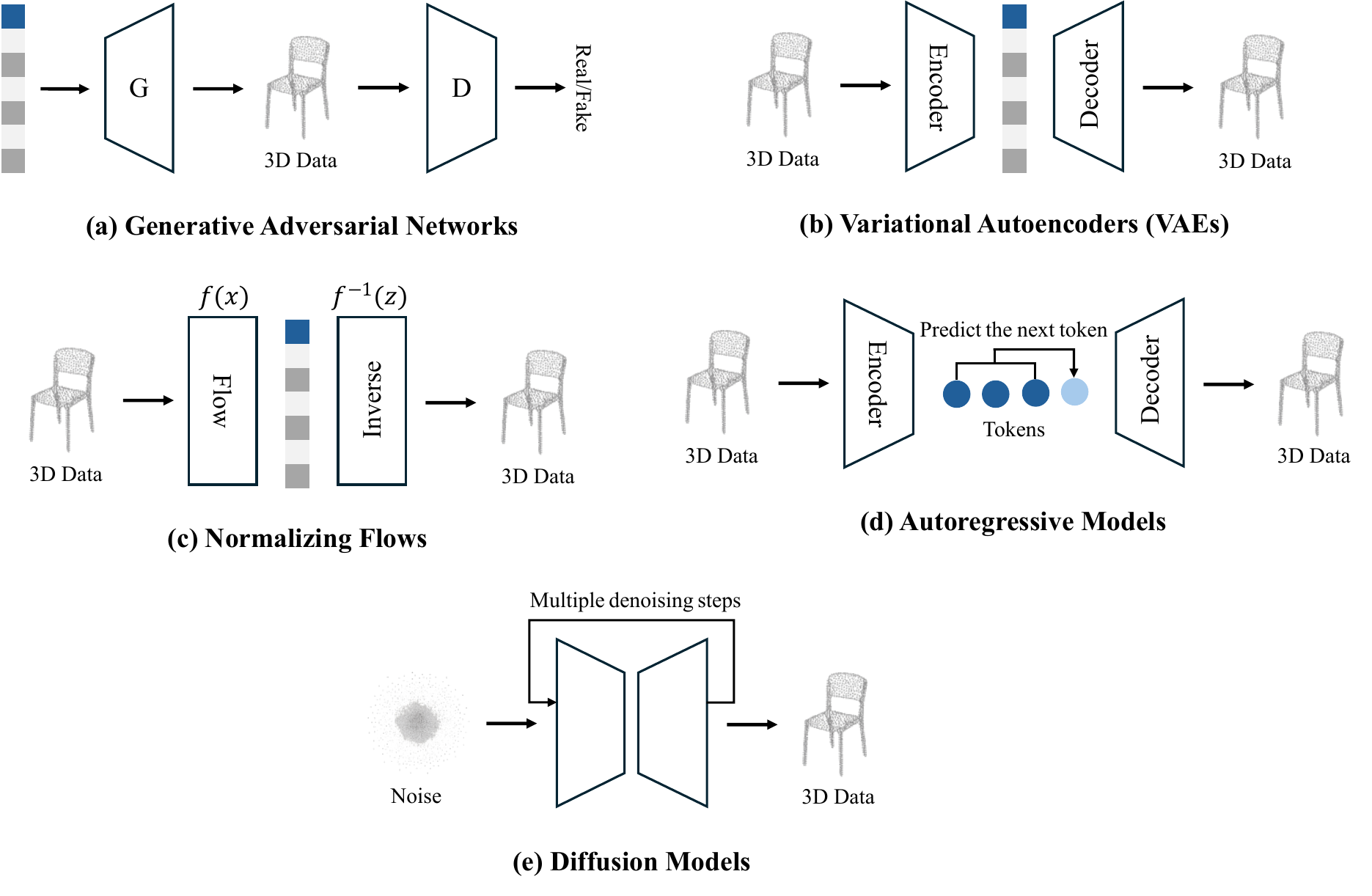}
\caption{Exemplary pipelines of feedforward 3D generation methods.}
\label{fig:exemplary_feeforward_gen_models}
\end{figure*}

Feedforward generation refers to the direct synthesis of 3D representations through generative models~\cite{lin_Kiss3DGenRepurposingImage_2025}. In this section, we explore these methods based on their generative models as shown in Figure \ref{fig:exemplary_feeforward_gen_models}, which include (a) Generative Adversarial Networks (GANs), (b) Variational Autoencoders (VAEs), (c) Normalizing Flows, (d) Autoregressive Models, and (e) Diffusion Models.

\subsubsection{Generative Adversarial Networks (GANs)}

Generative Adversarial Networks (GANs)~\cite{goodfellow_GenerativeAdversarialNets_2014} have demonstrated remarkable outcomes in the domain of 3D shape generation, consisting of a generator $G(\cdot)$ and a discriminator $D(\cdot)$, where the discriminator tries to classify real objects and objects synthesized by the generator, and the generator attempts to fool the discriminator~\cite{wu_LearningProbabilisticLatent_2016}.

The core idea is to integrate GANs with 3D shape representations (as discussed in Section \ref{section_3d_shape_representations}), enabling the generation of geometries such as point clouds (l-GAN/r-GAN~\cite{achlioptas_learning_2018}, tree-GAN~\cite{shu_treeGAN3DPointCloud_2019}, SP-GAN~\cite{li_SPGANSphereguided3D_2021}), voxel grids (3D-GAN~\cite{wu_LearningProbabilisticLatent_2016}, Z-GAN~\cite{knyaz_ImagetoVoxelModelTranslation_2019}), meshes (MeshGAN~\cite{cheng_MeshGANNonlinear3D_2019a}), or SDF (SurfGen~\cite{luo_SurfGenAdversarial3D_2021}, SDF-StyleGan~\cite{zheng_SDFStyleGANImplicitSDFBased_2022}). In this context, GANs serve as the underlying generative framework that learns to synthesize 3D shapes by mapping latent codes to complex geometric structures~\cite{li_SPGANSphereguided3D_2021}.

In the context of 3D shape generation, prior GAN-based methods such as 3D-GAN~\cite{wu_LearningProbabilisticLatent_2016}, l-GAN/r-GAN~\cite{achlioptas_learning_2018}, and Multi-chart Generation~\cite{ben-hamu_MultichartGenerativeSurface_2018} leverage explicit representations of real data to guide the training of generator networks. These approaches employ discriminators that assess 3D structures to supervise the generator, encouraging the synthesis of shapes that resemble the geometric realism of the training data. While these methods are capable of generating high-resolution shapes, they still struggle with thin structures and implausible objects. To address this limitation, SurfGen~\cite{luo_SurfGenAdversarial3D_2021} was introduced as an extension of implicit GANs with differentiable surface-aware rendering. SurfGen adopts DeepSDF~\cite{park_DeepSDFLearningContinuous_2019} as the generator within its proposed framework. To enable surface-level discrimination on objects represented as SDFs, the zero-level isosurface is first identified. Then, the marching cubes algorithm~\cite{lorensen_MarchingCubesHigh_1987} is applied to extract an explicit triangle mesh from the continuous SDF representation. For differentiable isosurface extraction, SurfGen leverages the MeshSDF method introduced in~\cite{remelli_MeshSDFDifferentiableIsoSurface_2020}.

Additional GAN-based methods have been proposed for 3D shape generation, each introducing distinct architectural innovations and geometric priors. One notable example is GraphCNN-GAN~\cite{valsesia_LearningLocalizedGenerative_2018}, which incorporates dynamic graph convolutions in the generator to better capture local geometric structures in point cloud generation, thereby enhancing the realism of fine details. Another approach, SP-GAN~\cite{li_SPGANSphereguided3D_2021}, introduces a spherical prior that guides the generator to produce points based on an underlying sphere parameterization. This encourages evenly distributed surface points and enables intuitive manipulation of shape parts, such as deforming a region by moving points along the sphere. MRGAN~\cite{gal_MRGANMultiRooted3D_2021} presents an unsupervised part-based generation framework, utilizing multiple latent roots in the generator. Each root is responsible for producing a component of the point cloud, achieving part-wise disentanglement. 

Similar to SurfGen, ShapeGAN~\cite{kleineberg_AdversarialGenerationContinuous_2020} employs a DeepSDF-style decoder as the generator and explores various 3D discriminators. This was one of the first GANs to operate in the learned implicit shape space, enabling the generation of higher-resolution geometry compared to voxel-based GANs.

\subsubsection{Variational Autoencoders (VAEs)}

Suppose we have a random variable $X$ for which we are building generative models. The variational autoencoder (VAE) is a framework that allows one to learn $P(X)$ from a dataset of observations of $X$~\cite{kingma_AutoEncodingVariationalBayes_2014, rezende_StochasticBackpropagationApproximate_2014, yang_pointflow_2019}. The VAE models the data distribution via a latent variable $z$ with a prior distribution $P_\psi(z)$, and a decoder $P_\theta(X\,|\,z)$ which captures the (hopefully simpler) distribution of $X$ given $z$. During training, it additionally learns an inference model (or encoder) $Q_\phi(z\,|\,X)$. The encoder and decoder are jointly trained to maximize a lower bound on the log-likelihood of the observations

\begin{align}
\log P_\theta(X) &\geq \log P_\theta(X) - D_{KL}(Q_\phi(z\,|\,X)\,\|\,P_\theta(z\,|\,X)) \notag \\
&= \mathbb{E}_{Q_\phi(z\,|\,X)}[\log P_\theta(X\,|\,z)] - D_{KL}(Q_\phi(z\,|\,X)\,\|\,P_\psi(z)) \notag \\
&\triangleq \mathcal{L}(X;\, \phi,\, \psi,\, \theta)
\end{align}

which is also called the evidence lower bound (ELBO). The ELBO can be interpreted as the sum of the negative reconstruction error (the first term) and a latent space regularizer (the second term). In practice, $Q_\phi(z\,|\,X)$ is usually modeled as a diagonal Gaussian $\mathcal{N}(z\,|\,\mu_\phi(X),\, \sigma_\phi(X))$ whose mean and standard deviation are predicted by a neural network with parameters $\phi$. To efficiently optimize the ELBO, sampling from $Q_\phi(z\,|\,X)$ is done by reparametrizing $z$ as $z = \mu_\phi(X) + \sigma_\phi(X) \, \cdot \epsilon$ where $e \thicksim \mathcal{N}(0,I)$.

In this context, VAEs~\cite{kingma_AutoEncodingVariationalBayes_2014} are probabilistic generative models composed of two neural networks: an encoder and a decoder. The encoder maps an input data point to a latent space by producing the parameters of a variational distribution--tipycally a multivariate Gaussian--from which latent vectors can be sampled. This enables the encoder to generate multiple samples that originate from the same underlying distribution. The decoder then maps these latent variables back to the data space, reconstructing or generating new samples. Both networks are typically trained jointly using the reparametrization trick, which allows for gradient-based optimization through the sampling process. Although the noise model's variance can also be learned, it is sometimes treated separately for stability or modeling purposes~\cite{li_Advances3DGeneration_2024}.

\cite{brock_GenerativeDiscriminativeVoxel_2016} introduced a voxel-based VAE alongside a graphical user interface for exploring the latent space of 3D generative models. Their framework also includes a deep voxel-based convolutional neural network for classification. In parallel, SDM-Net~\cite{gao_SDMNETDeepGenerative_2019} focuses on generating structured meshes composed of deformable parts. The method uses one VAE network to model parts and another to model the whole object. Building upon this idea, TM-Net~\cite{gao_TMNETDeepGenerative_2021} extends the approach to generate texture maps for meshes in a part-aware manner.

In the context of 3D shape generation, GRASS~\cite{li_GRASSGenerativeRecursive_2017} proposes a structure-aware generation framework, which learns a latent space of shape layouts (parts and their connections) using a recursive decoder that splits shape parts hierarchically. GRASS was the first to generate 3D shapes as a composition of parts, producing plausible novel combinations (e.g., new chair designs) and enabling interpolation at the part-structure level. Subsequently, ShapeVAE~\cite{nash_ShapeVAEShapeVariationalAutoencoder_2017} was one of the first VAE-based models proposed for 3D shape generation. It introduced a probabilistic latent space over 3D object geometry, enabling the generation of novel shape variants through latent sampling. However, a key limitation of the model was its low output resolution. ShapeVAE also incorporated part segmentation into its latent representation, allowing for intuitive shape analogies and part-level manipulations. 

Additional VAE-based models have since been proposed, including AtlasNet~\cite{groueix_AtlasNetPapierMacheApproach_2018}, which reconstructs surfaces from learned parametrizations; SetVAE~\cite{kim_SetVAELearningHierarchical_2021}, which extends the VAE framework to handle unordered point sets using hierarchical structure-aware encoding; Michelangelo~\cite{zhao_MichelangeloConditional3D_2023}, which leverages part-aware priors in conditional generative tasks; TAR3D \cite{zhang_TAR3DCreatingHighQuality_2025}, which extends the learning capabilities of the next-token prediction paradigm to conditional 3D object generation, enabling more effective sequence-based modeling of 3D structures; and COD-VAE \cite{cho_Representing3DShapes_2025}, which introduces a representation framework that encondes 3D shapes into a set of compact 1D latent vectors, providing a lightweight yet expressive latent space for generation and reconstruction.

In synthesis, VAEs are often used in combination with other generative paradigms to enhance structure modeling and multimodal conditioning. Compared to GAN-based approaches, the training of VAEs is considerably more stable than GANs. However, VAEs tend to produce more blurred results compared to GANs~\cite{li_Advances3DGeneration_2024}. 


\subsubsection{Normalizing Flows}

A normalizing flow~\cite{rezende_VariationalInferenceNormalizing_2015} is a series of invertible mappings that transform an initial known distribution $p_Z(z)$ to a complex data distribution $p_X(x)$~\cite{kim_SoftFlowProbabilisticFramework_2020}.

In the context of 3D shape generation, PointFlow~\cite{yang_pointflow_2019} learns the distribution of shapes, each shape itself being a distribution of points using continuous normalizing flows~\cite{chen_NeuralOrdinaryDifferential_2018, grathwohl_FFJORDFreeFormContinuous_2018}. PointFlow was the first to show that flow-based models can generate 3D point sets with improved coverage (diversity), though at a higher computational cost. To address this limitation, the Discrete Point Flow (DPF) networks~\cite{klokov_DiscretePointFlow_2020} were introduced, which use affine coupling layers (a discrete flow) to map noise to point cloud coordinates. DPF networks achieve a significant speed-up in generation with comparable fidelity, making flow models more practical for 3D.

SoftFlow~\cite{kim_SoftFlowProbabilisticFramework_2020} is a framework for training normalizing flows on the manifold. In order to bridge the gap between the dimensions of the data and the target latent variable, the authors propose a method to estimate a conditional distribution of the perturbed data. SoftFlow resolves the dimension mismatch and improves the stability of flow-based 3D shape generation.

Subsequent works--such as Point Straight Flow (PSF)~\cite{wu_FastPointCloud_2023} and EAGLE~\cite{wang_EAGLEContextualPoint_2025}--have further advanced the use of normalizing flows for 3D shape generation. In particular, EAGLE proposes adaptive continuous normalizing flows by introducing a control mechanism with adaptive learning rates to the bias term.

\subsubsection{Autoregressive Models}

Flow-based models and auto-regressive models can both perform exact likelihood evaluation, while flow-based models are much more efficient to sample from~\cite{yang_pointflow_2019}.

Autoregressive networks model current values as a function of their own previous values~\cite{sun_PointGrowAutoregressivelyLearned_2020}. According to~\cite{li_Advances3DGeneration_2024}, a 3D object can be represented as a joint probability of the occurrences of multiple 3D elements:

\begin{equation}
p(x_0,\, x_1,\, \ldots,\, x_n),
\end{equation}

where $x_i$ denotes the $i$-th element, such as a point or voxel coordinate. Estimating this joint probability becomes increasingly complex as the number of variables grows. To make the problem more tractable, it can be factorized as a product of conditional probabilities:

\begin{equation}
p(x_0,\, x_1,\, \ldots,\, x_n) = p(x_0) \prod_{i=1}^{n} p(x_i \mid x_{<i}),
\end{equation}

which enables the learning of conditional distributions and allows joint probability estimation via sampling.

Autoregressive models for data generation follow this formulation by modeling each element's probability conditioned on the preceding ones. Given an ordered sequence of elements $x_0,\, x_1,\, \ldots,\, x_n$, a model can be trained to predict each $x_i$ based on its predecessors $x_0,\, \ldots,\, x_{i-1}$:

\begin{equation}
p(x_i \mid x_{<i}) = f(x_0,\, \ldots,\, x_{i-1}),
\end{equation}

where $f$ denotes the model that learns the conditional probability distribution. This training process is commonly referred to as \emph{teacher forcing}. Once trained, the model can be employed in an autoregressive manner to sequentially generate each element step by step:

\begin{equation}
x_i = \arg\max p(x_i \mid x_{<i}).
\end{equation}

In the context of 3D shape generation, several studies have explored autoregressive modeling as a generative approach. PointGrow~\cite{sun_PointGrowAutoregressivelyLearned_2020}, in turn, employs a Transformer-like self-attention mechanism to autoregressively predict points step by step, each conditioned on the previously generated points. It captures long-range dependencies and yields diverse shapes, but is inherently slower to sample. Inspired by the autoregressive design of PointGrow, PolyGen~\cite{nash_PolyGenAutoregressiveGenerative_2020} introduces a transformer-based framework for generating 3D meshes. It employs two sequential transformer networks: a vertex model, which unconditionally models mesh vertices, and a face model, which models the mesh faces conditioned on input vertices. This approach enables the structured synthesis of meshes while maintaining geometric consistency through autoregressive modeling.

Subsequent works have extended autoregressive modeling to various 3D representations. AutoSDF~\cite{mittal_AutoSDFShapePriors_2022} generates 3D shapes represented by volumetric truncated-signed distance functions (T-SDF). It employs a VQ-VAE \cite{vandenoord_NeuralDiscreteRepresentation_2017} framework to learn a quantized codebook that captures local geometric patterns within T-SDFs. Shapes are encoded as discrete tokens from this codebook and modeled using a transformer-based architecture in a non-sequential autoregressive manner. Also, AutoSDF is capable of completing shapes based on images or text. 

Additional approaches include GraphRNN~\cite{you_GraphRNNGeneratingRealistic_2018}, ScanComplete~\cite{dai_ScanCompleteLargeScaleScene_2018}, StructureNet~\cite{mo_StructureNetHierarchicalGraph_2019}, ShapeFormer~\cite{yan_ShapeFormerTransformerbasedShape_2022}, ShapeCrafter~\cite{fu_ShapeCrafterRecursiveTextConditioned_2023}, Octree Transformer~\cite{ibing_OctreeTransformerAutoregressive_2023}, and PointARU~\cite{meng_PointARU3DPoint_2025}, which adopts a coarse-to-fine autoregressive up-sampling approach, eliminating the need to define a fixed generation order for point sequences. In addition, ShapeGPT \cite{yin_ShapeGPT3DShape_2025} introduces a multi-modal framework that leverages large pre-trained language models to address multiple tasks, including text-to-shape, shape-to-text, shape completion, and shape editing.

These methods explore diverse input structures--such as graphs, hierarchical trees, and octrees--while leveraging autoregressive priors to generate semantically and geometrically coherent shapes.

\subsubsection{Diffusion Models}

Diffusion models~\cite{sohl_dickstein_DeepUnsupervisedLearning_2015, ho_DenoisingDiffusionProbabilistic_2020} are a class of generative models that convert simple known distributions (e.g., a Gaussian) into complex data distributions. These models generate data by simulating a reverse diffusion process, gradually transforming a noise sample into a structured output. Their application to 3D shape generation has shown promising performance in capturing fine geometric details and enabling flexible conditioning mechanisms~\cite{rombach_HighResolutionImageSynthesis_2022, li_Advances3DGeneration_2024}.

Diffusion models, particularly the Denoising Diffusion Probabilistic Models (DDPMs)~\cite{ho_DenoisingDiffusionProbabilistic_2020}, define a Markovian forward process $q (x_t \mid x_{t-1})$ that progressively adds Gaussian noise to the data: 

\begin{equation} q(x_t \mid x_{t-1}) = \mathcal{N}(x_t;\, \sqrt{1 - \beta_t}\, x_{t-1},\, \beta_t \, \mathbf{I}), \end{equation}

where $x_0 \sim q(x_0)$ is the real data distribution, $t \in \{1,\,\dots,\,T\}$ indexes time steps, and $\beta_t$ is a variance schedule. After enough steps, $x_T \sim \mathcal{N}(0,\mathbf{I})$ approximates pure Gaussian noise.

The reverse process is parameterized by a neural network $p_\theta(x_{t-1} \mid x_t)$ that learns to denoise $x_t$ step by step:

\begin{equation} p_\theta(x_{t-1} \mid x_t) = \mathcal{N}(x_{t-1};\, \mu_\theta(x_t,\, t),\, \Sigma_\theta(x_t,\, t)). \end{equation}

The model is trained to predict either the original data $x_\theta$ or the noise $\epsilon$ (depending on the parameterization) by minimizing a variational bound or simplified noise prediction loss.

In the 3D domain, diffusion models have been adapted to generate a wide range of shape representations, including point clouds, meshes, voxel grids, point-voxel, and neural fields.

Several methods have explored diffusion directly in the point cloud space. For instance, ShapeGF
\cite{cai_ShapeGFLearningGradient_2020} builds upon a denoising score-matching framework to learn the underlying data distributions of point clouds. PVD~\cite{zhou_3DShapeGeneration_2021} introduces a bidirectional diffusion model that learns mappings between point clouds and voxel grids, enabling both generation and reconstruction tasks. Similarly, DPM~\cite{luo_DiffusionProbabilisticModels_2021} learns a noise-conditioned score function to refine noisy point clouds into clean samples progressively. Building on these foundations, LION~\cite{zeng_LIONLatentPoint_2022} introduces a latent diffusion model specifically tailored for point cloud generation. By operating in a learned latent space, LION improves sampling efficiency and scalability while maintaining geometric fidelity. TIGER~\cite{ren_TIGERTimeVaryingDenoising_2024} further advances this direction by incorporating a time-varying denoising process in a single triangle-grid representation, achieving high-resolution shape synthesis with fine topological structure.

Beyond point clouds, MeshDiffusion~\cite{liu_MeshDiffusionScorebasedGenerative_2022} and SLIDE~\cite{lyu_SLIDEControllableMeshGeneration_2023} employ score-based and controllable diffusion strategies, respectively, to generate meshes directly with controllable topology and attributes. Tetrahedral Diffusion Models~\cite{kalischek_TetraDiffusionTetrahedralDiffusion_2025} explore volumetric mesh generation using tetrahedral elements, enabling continuous and differentiable surface extraction.

Additional works have extended diffusion to hybrid and latent representations. Shap-E~\cite{jun_ShapEGeneratingConditional_2023} learns a unified latent field for conditional generation from images to text. Wavelet Diffusion~\cite{hu_NeuralWaveletdomainDiffusion_2024} and WaLa~\cite{sanghi_WaveletLatentDiffusion_2024} introduce wavelet-based latent diffusion to improve resolution and multi-scale control. SALAD~\cite{koo_SALADPartLevelLatent_2023} and CoPart \cite{dong_OneMoreContextual_2025} focus on part-level generation using latent diffusion models that can synthesize individual object components. Point-E~\cite{nichol_PointESystemGenerating_2022} combines the power of GLIDE~\cite{nichol_GLIDEPhotorealisticImage_2022} for conditional image generation with a diffusion model-based decoder that produces point clouds from synthesized views. XCube~\cite{ren_XCubeLargeScale3D_2024} introduces a scalable approach for 3D shape generation based on sparse voxel hierarchies, enabling memory-efficient synthesis of high-resolution geometry across large and diverse object categories.

Recent advances also include models that further expand the capabilities of latent diffusion for 3D shape generation. GPLD3D~\cite{dong_GPLD3DLatentDiffusion_2024} proposes latent diffusion with enforced geometric and physical priors to improve the realism and plausibility of generated shapes. The triplane-based 3D-aware Diffusion model with TransFormer (DiffTF)~\cite{cao_LargeVocabulary3DDiffusion_2023} introduces a transformer-based latent diffusion framework capable of generating shapes across a wide range of object categories. Sin3DM~\cite{wu_Sin3DMLearningDiffusion_2023} focuses on single-shot learning, training a diffusion model from just a single 3D textured shape. Finally, Direct3D~\cite{wu_Direct3DScalableImageto3D_2024} presents a scalable image-to-3D generation approach based on a 3D latent diffusion transformer, enabling efficient and high-quality synthesis conditioned on images. Likewise, Hi3DGen \cite{ye_Hi3DGenHighfidelity3D_2025} introduces a normal bridging strategy that further enhances fidelity and efficiency in image-conditioned 3D generation. 

Together, these approaches showcase the versatility and scalability of diffusion models for handling diverse 3D shape representations, including explicit, implicit, and hybrid representations. As research in this area continues to advance, it is expected that diffusion models will play a crucial role in pushing the boundaries of 3D shape generation across a wide range of applications~\cite{li_Advances3DGeneration_2024}.

\section{Open-Source Datasets for 3D Generation}\label{sec:datasets}

In the context of 3D shape generation, data plays a critical role in guiding both the learning and synthesis processes. Using an appropriate dataset can help the learning and generation process based on the application~\cite{xu_SurveyDeepLearningbased_2023}.

However, no single dataset satisfies all the requirements across different shape representations. The shapes in the dataset need to be preprocessed (e.g., voxelized, sampled) in order to be used for different representations~\cite{xu_SurveyDeepLearningbased_2023}.

Early 3D shape datasets were introduced with the aim of supporting tasks such as shape classification, recognition, and object modeling.  One of the earliest is the Princeton Shape Benchmark~\cite{shilane_PrincetonShapeBenchmark_2004a}, which includes approximately 1800 polygonal models collected from the World Wide Web. The KIT object models database~\cite{kasper_KITObjectModels_2012b} employs a specialized acquisition rig consisting of a 3D digitizer, a turntable, and stereo cameras mounted on a sled that can move along a bent rail to capture 3D geometric data of everyday objects. Over 100 objects have been digitized using this system.

To address fine-pose estimation of objects in 2D images given a 3D model, the dataset introduced in ~\cite{lim_ParsingIKEAObjects_2013}, named as IKEA objects, includes 800 images and 225 IKEA-style CAD models sourced from Google Warehouse. Other datasets have been introduced for specific applications, including robotic manipulation~\cite{calli_BenchmarkingManipulationResearch_2015, morrison_EGADEvolvedGrasping_2020}, 3D shape retrieval (Shrec’14~\cite{godil_Shrec14TrackLarge_2014}), single-image 3D shape reconstruction (Pix3D~\cite{sun_Pix3DDatasetMethods_2018}), and scanned objects (Google Scanned Objects~\cite{downs_GoogleScannedObjects_2022}). The BigBird dataset~\cite{singh_BigBIRDLargescale3D_2014} contributes a large-scale collection of 3D object instances along with multi-view RGB images, depth maps, camera pose annotations, and object segmentations.

\begin{figure*}[tbp]
\centering
\includegraphics[width=1\textwidth]{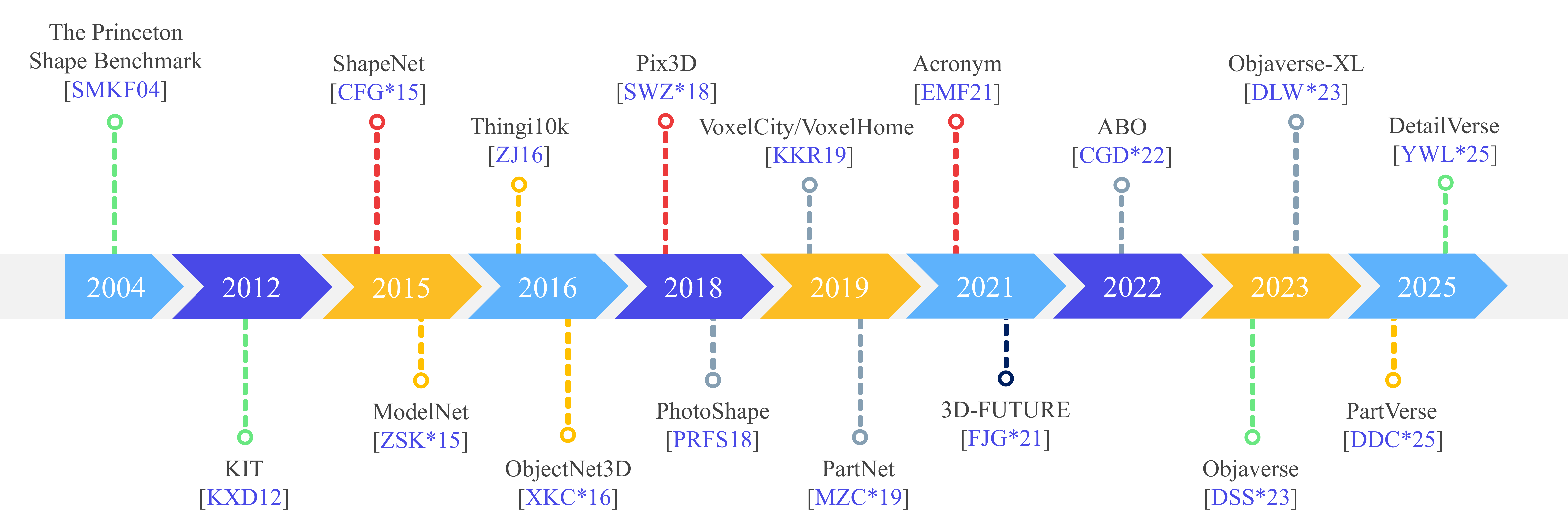}
\caption{Timeline of Open-Source Datasets for 3D Generation.}
\label{fig:timeline_datasets}
\end{figure*}

Despite their historical significance, these datasets are very small--they only contain thousands of objects--and are not widely adopted in current 3D shape generation research. Figure \ref{fig:timeline_datasets} shows a timeline of widely used open-source datasets for 3D shape generation. In the remainder of this section, we describe these datasets based on the state-of-the-art methods described in Section \ref{3d_generation_methods}.

One of the most widely used datasets in 3D generation is ShapeNet~\cite{chang_ShapeNetInformationRich3D_2015}, which includes two main subsets: ShapeNetCore and ShapeNetSem. The full dataset comprises nearly 3 million textured 3D computer-aided design (CAD) models, spanning 3,135 object categories. ShapeNetCore, the most commonly used subset, contains approximately 51,300 models across 55 object categories, each with manually verified annotations for category labels and object alignment. The standard experimental setup for evaluating generative models on ShapeNet is often adapted from the benchmark introduced by PointFlow.


ModelNet~\cite{zhirongwu_3DShapeNetsDeep_2015} is another well-known, extensive shape dataset of 3D CAD models. It is one of the most widely used benchmarks for point cloud analysis. It contains 151,128 objects belonging to 660 categories.

ObjectNet3D~\cite{xiang_ObjectNet3DLargeScale_2016} contains 3D shapes and 2D images. The 2D images are aligned with the corresponding 3D shapes. ObjectNet consists of 100 categories, 90,127 images, 201,88 objects in these images, and 44,147 3D shapes. The objects in the 2D images are aligned with the 3D shapes, and the alignment provides both accurate 3D pose annotation and the closest 3D shape annotation for each 2D object.

Thingi10k~\cite{zhou_Thingi10KDataset10000_2016} collects 10,000 3D printing models from the Thingiverse repository. Thingi10k covers many categories and provides a balanced representation of real-world data regarding mesh complexity and quality.

PhotoShape~\cite{park_PhotoShapePhotorealisticMaterials_2018a} produces 29,133 synthetic relightable chair objects with photorealistic materials, based on online data.

In~\cite{knyaz_ImagetoVoxelModelTranslation_2019}, the authors propose two novel datasets, VoxelCity and VoxelHome, designed to facilitate research in image-to-voxel translation. The VoxelCity dataset includes 3D models of 21 urban scenes, including 18,836 RGB images with reconstructed 3D models, ground-truth 3D CAD models, depth maps, and 6D poses of seven object classes. The VoxelHome dataset presents 3D models of 7 indoor scenes, composed of 17,580 color images with reconstructed 3D models, ground-truth 3D CAD models, depth maps, and 6D poses of nine object classes.

PartNet~\cite{mo_PartNetLargeScaleBenchmark_2019} is a consistent, large-scale dataset on top of ShapeNet, with fine-grained, hierarchical, instance-level 3D part information. PartNet comprises 573,585 part instances over 26,671 3D models covering 24 object categories. It is commonly used for point cloud-related tasks.

Acronym~\cite{eppner_ACRONYMLargeScaleGrasp_2021} provides watertight mesh models spanning 262 categories, derived from the original ShapeNet dataset, making it easy to convert them to other representations, e.g., SDF.

3D-FUTURE~\cite{fu_3DFUTURE3DFurniture_2021d} is a large-scale dataset that contains 20,240 realistic indoor images and the associated 9,992 unique 3D furniture models with rich geometry details and high-resolution textures. 3D-FUTURE also offers instance segmentation annotation and image rendering information, including six degrees of freedom (6DoF).

The Amazon Berkeley Objects (ABO)~\cite{collins_ABODatasetBenchmarks_2022} dataset was designed to help bridge the gap between real and virtual 3D worlds. ABO is derived from Amazon.com product listings. It contains 147,702 product listings associated with 398,212 unique catalog images. ABO also includes 7,953 products with corresponding artist-designed 3D meshes.

Objaverse~\cite{deitke_ObjaverseUniverseAnnotated_2023} is a large-scale, high-quality dataset of 3D objects with rich annotations, including natural language captions. Objaverse contains over 800K textured 3D models spanning a wide variety of categories and use cases. Its successor, Objaverse-XL~\cite{deitke_ObjaverseXLUniverse10M_2023} further extends Objaverse to a larger 3D dataset of 10.2M 3D objects, establishing it as one of the largest publicly available corpora for multimodal 3D learning.

More recently, \cite{dong_OneMoreContextual_2025} introduces PartVerse, a large-scale 3D part dataset comprising 91k high-quality parts from 12k objects with detailed descriptions. PartVerse is a curated and annotated dataset from Objaverse, and exhibits enhanced diversity in 175 categories and includes realistic textures, making it a valuable benchmark for part-level 3D understanding and generation.

Likewise, \cite{ye_Hi3DGenHighfidelity3D_2025} proposes DetailVerse, a dataset of 700k high-quality meshes, synthesized through a Text$\rightarrow$Image$\rightarrow$3D generation pipeline. This large-scale collection offers fine-grained geometric detail and high-fidelity assets.

These large-scale 3D datasets have the potential to facilitate large-scale training and boost the performance of 3D generation~\cite{li_Advances3DGeneration_2024}, specifically in the context of 3D shape generation.

\section{Metrics}\label{sec:metrics}

Evaluating the sample quality of generative models remains a challenging task~\cite{you_GraphRNNGeneratingRealistic_2018}, particularly in the context of 3D shape generation, where no single metric can reliably quantify how “good” a generated shape is~\cite{wang_DiffusionModels3D_2025}.

In this section, we review the most commonly used metrics for evaluating 3D shape generation methods. We organize these metrics according to the specific evaluation objectives they address, and we focus on those metrics that are widely adopted in state-of-the-art literature.

\subsection{Fidelity Metrics}\label{sec:fidelity_metrics}

Fidelity metrics quantify the geometric similarity between a generated shape and a reference ground-truth shape. These metrics are essential in tasks such as shape synthesis, reconstruction, competition, or the evaluation of individual shapes.

\subsubsection{Chamfer Distance (CD)}

Chamfer Distance was originally introduced in the context of computer vision by~\cite{barrow_recovering_1978}, to compare object contours and edge maps. The modern formulation computes the average distance from each point in one point cloud ($P$) to its nearest neighbor in the second point cloud ($Q$), and vice versa.

Popularized in 3D point cloud generation by~\cite{fan_point_2016}, CD has since become a standard metric for evaluating geometric fidelity between a generated shape and a ground truth shape. It is used extensively in 3D reconstruction, shape completion, and point set generation, as well as in higher-level metrics such as 1-NNA, MMD, and Coverage that rely on pairwise distances.

\cite{fan_point_2016, aguirre_DatasetfreeApproachSelfsupervised_2025} defines the Chamfer Distance between two point clouds $S_1$ and $S_2  \subseteq \mathbb{R}^3$ as:

\begin{equation}
{CD}(S_1, S_2) =  \frac{1}{|S_1|} \sum_{x\, \in\, S_1} \min_{y\, \in\, S_2} \|x - y\|_2^2 + \frac{1}{|S_2|} \sum_{y\, \in\, S_2} \min_{x\, \in\, S_1} \|x - y\|_2^2
\end{equation}

In the strict sense, ${CD}(\cdot, \cdot)$ is not a distance function because the triangle inequality does not hold. For each point, the algorithm of CD finds the nearest neighbor in the other set and sums the squared distances. The lower, the better.

\subsubsection{Earth Mover´s Distance (EMD)}

\cite{rubner_earth_2000} formalized EMD as a metric for comparing feature distributions in image retrieval tasks based on optimal transport theory. EMD measures the minimum cost required to transform one distribution into another, considering one-to-one assignments between elements.

In~\cite{achlioptas_learning_2018}, EMD was used to evaluate the quality of 3D point clouds generated by generative models by comparing the distribution of generated points to that of real data. EMD is also employed as a base distance in set-level evaluation, such as 1-NNA, MMD, and Coverage, offering a stricter measure than CD by enforcing globally optimal point-wise assignments. Consider $S_1, S_2  \subseteq \mathbb{R}^3$, the Earth Mover's Distance is defined as:

\begin{equation}
{EMD}(S_1, S_2) = \min_{\phi :\, S_1\, \to\, S_2} \sum_{x\, \in\, S_1} \|x - \phi(x)\|_2^2
\end{equation}

Measures the optimal cost of transforming one point cloud into another by considering the global structure of the distributions. The EMD distance solves an optimization problem, namely, \emph{the assignment problem}. For all but a zero-measure subset of point set pairs, the optimal bijection $\phi$ is unique and invariant under infinitesimal movement of the points~\cite{fan_point_2016}. Although EMD is computationally more expensive than Chamfer Distance, it provides a more accurate and globally consistent measure of geometric similarity.

\subsubsection{F-score}

Introduced in information retrieval by~\cite{rijsbergen_InformationRetrieval_1979} as the harmonic mean of precision and recall.

Used in 3D reconstruction benchmarks to combine accuracy and completeness. For example,~\cite{knapitsch_tanks_2017} employed the F-Score (at a distance threshold) to evaluate reconstructed point clouds, making it a standard for comparing 3D reconstruction quality. Let $\mathcal{G}$ denote the ground truth shape and $\mathcal{R}$ the reconstructed point cloud. For a reconstructed point $r \in \mathcal{R}$, its closest distance to the ground truth is defined as:

\begin{equation}
e_{r\, \rightarrow\, \mathcal{G}} = \min_{g\, \in\, \mathcal{G}} \| r - g \|.
\end{equation}

This value can be used to compute the \emph{precision} at a given threshold $d$:

\begin{equation}
P(d) = \frac{100}{|\mathcal{R}|} \sum_{r\, \in\, \mathcal{R}} [e_{r\, \rightarrow\, \mathcal{G}} < d],
\end{equation}

where $[\cdot]$ denotes the Iverson bracket, returning 1 when the condition is true and 0 otherwise. $P(d)$ ranges from 0 to 100 and is interpreted as the percentage of reconstructed points that lie within a distance $d$ from the ground truth.
Analogously, for each ground truth point $g \in \mathcal{G}$, its distance to the closest reconstructed point is:

\begin{equation}
e_{g\, \rightarrow\, \mathcal{R}} = \min_{r\, \in\, \mathcal{R}} \| g - r \|.
\end{equation}

This leads to the definition of \emph{recall} at threshold $d$:

\begin{equation}
R(d) = \frac{100}{|\mathcal{G}|} \sum_{g\, \in\, \mathcal{G}} [e_{g\, \rightarrow\, \mathcal{R}} < d].
\end{equation}

The final \emph{F-score} combines both quantities as their harmonic mean:

\begin{equation}
F(d) = \frac{2 P(d)\, R(d)}{P(d) + R(d)}.
\end{equation}

The F-score provides a robust summary of how well the reconstruction captures both the overall geometry (recall) and the surface accuracy (precision). It has the desirable property that if either $P(d) \rightarrow 0$ or $R(d) \rightarrow 0$, then $F(d) \rightarrow 0$, unlike the arithmetic mean. 

In the context of point clouds, a high F-score indicates that most points are well aligned with respect to the ground truth.

\subsubsection{Intersection over Union (IoU)}

Proposed by~\cite{jaccard_1901} for comparing set similarity in ecology. Also known as the Jaccard Index, it measures the overlap between two sets as a ratio of intersection over union.

First applied to 3D model evaluation in voxel-based reconstructions. For instance, 3D-R2N2~\cite{choy_3d-r2n2_2016} reported IoU between predicted and ground-truth voxels to assess single-view 3D reconstruction performance. IoU has since been ubiquitous for evaluating 3D shape prediction accuracy.

3D-R2N2 used IoU to evaluate a 3D voxel reconstruction and its ground truth voxelized model. More formally, 

\begin{equation}
\textit{IoU} = 
\frac{
    \sum_{i,\,j,\,k} \left[ I(p_{(i,\,j,\,k)} > t) \, I(y_{(i,\,j,\,k)}) \right]
}{
    \sum_{i,\,j,\,k} \left[ I(I(p_{(i,\,j,\,k)} > t) + I(y_{(i,\,j,\,k)})) \right]
}
\end{equation}

where $I(\cdot)$ is an indicator function and $t$ is a voxelization threshold. Higher IoU indicates better reconstructions.

\subsection{Diversity and Distribution Metrics}

These metrics evaluate statistical properties of the set of generated shapes in comparison to the set of original shapes. They are essential in unconditional generation tasks, where no specific target shape is expected as output.

\subsubsection{Minimum Matching Distance (MMD)}

Introduced by~\cite{achlioptas_learning_2018} specifically for evaluating generative models of 3D point clouds. It is defined as the average nearest-neighbor distance from each reference shape to the closest generated shape (using a chosen point-set distance). MMD quantifies fidelity (lower is better). 

Although this metric technically measures distance like fidelity metrics, it is grouped here because its predominant use in the literature is to compare sets of generated and real shapes. For example, PointFlow reports MMD (using CD and EMD) to evaluate differences between point cloud distributions, establishing MMD as a standard metric for assessing the diversity-quality trade-off in 3D shape generation.

For each point cloud in the reference set, the distance to its nearest neighbor in the generated set is computed and averaged~\cite{yang_pointflow_2019}:

\begin{equation}
\text{MMD}(S_g,\, S_r) = \frac{1}{|S_r|} \sum_{Y\, \in\, S_r} \min_{X\, \in\, S_g} D(X,\, Y),
\end{equation}

where $D(\cdot,\cdot)$ can be either CD or EMD. However, MMD is actually very insensitive to low-quality point clouds in $S_g$, since they are unlikely to be matched to real point clouds in $S_r$. From this point forward, we will refer to $S_g$ as the set of ground-truth objects, and $S_r$ as the set of generated objects.

\subsubsection{Coverage (COV)}

Also proposed in~\cite{achlioptas_learning_2018} as a complement to MMD, COV measures the fraction of real samples that have a generated sample within some distance threshold (or that are the nearest neighbor match). It reflects how well the generated set covers the models of the target distribution. Higher is better.

Used from~\cite{achlioptas_learning_2018} onward. For instance, PointFlow evaluates COV (\%) to ensure the generative model captures diverse modes of ShapeNet shapes. COV, together with MMD, has become standard for assessing diversity vs. fidelity in 3D point cloud generation.

For each point cloud in the generated set, its nearest neighbor in the reference set is marked as a match~\cite{yang_pointflow_2019}:

\begin{equation}
\text{COV}(S_g,\, S_r) = \frac{\left| \left\{ \arg\min_{Y \,\in\, S_r} D(X, \,Y) \mid X \in S_g \right\} \right|}{|S_r|},
\end{equation}

where $D(\cdot,\cdot)$ can be either CD or EMD.

\subsubsection{1-Nearest Neighbor Accuracy (1-NNA)}

Introduced by~\cite{lopez-paz_revisiting_2017} as a classifier-based two-sample test. The idea is to pool real and generated samples and use a 1-NN classifier (leave-one-out) to identify each sample's source; the accuracy of this test indicates if the two distributions are identical (50\% is ideal for identical distributions).

Applied later by PointFlow to evaluate 3D generative models. 1-NN accuracy (denoted 1-NNA) is used to directly measure distributional similarity between generated and real 3D point sets. PointFlow showed that 1-NN is a sensitive metric, giving lower (closer to 50\%) values when the generated shapes better approximate the real shape distribution. Let $S_{-X} = S_r \cup S_g - {X}$ and $N_X$ be the nearest neighbor of $X$ in $S_{-X}$. 1-NNA is the leave-one-out accuracy of the 1-NN classifier:

\begin{equation}
\text{1-NNA}(S_g,\, S_r) = 
\frac{
\sum_{X\, \in\, S_g} \mathbb{I}[N_X \in S_g] + 
\sum_{Y\, \in\, S_r} \mathbb{I}[N_Y \in S_r]
}{
|S_g| + |S_r|
},
\end{equation}

where $\mathbb{I}[\cdot]$ is the indicator function. For each sample, the 1-NN classifier classifies it as coming from $S_r$ or $S_g$ according to the label of its nearest sample.

According to~\cite{yang_pointflow_2019}, unlike COV and MMD, 1-NNA directly measures distributional similarity and takes both diversity and quality into account.

\subsubsection{Jensen-Shannon Divergence (JSD)}

Formally defined by~\cite{lin_divergence_1991} as a symmetrized and smoothed version of Kullback-Leibler divergence. It measures similarity between two probability distributions, with 0 meaning identical distributions.

\cite{achlioptas_learning_2018} introduced JSD to 3D generation by transforming point clouds into a voxel occupancy grid and computing the divergence between occupancy probabilities of real vs. generated sets. This became an early criterion for generative point cloud evaluation. Given two point cloud sets $A$ and $B$, the JSD measures the similarity between their voxel-based occupancy distributions $P_A$ and $P_B$ as follows:

\begin{equation}
\text{JSD}(P_A \parallel P_B) = \frac{1}{2} D(P_A \parallel M) + \frac{1}{2} D(P_B \parallel M),
\end{equation}

where $M = \frac{1}{2}(P_A + P_B)$, and $D(\cdot \parallel \cdot)$ is the KL-divergence between the two distributions~\cite{kullback_InformationSufficiency_1951}.

\subsubsection{Single Shape Fréchet Inception Distance (SSFID)}

Based on the Fréchet Inception Distance (FID) proposed by~\cite{heusel_gans_2017} for images. FID measures the distance between feature distributions of real vs. generated data (using Inception-v3 features) and has been widely adopted for GAN evaluation.

To extend this idea to 3D, \cite{wu_LearningGenerate3D_2022a} coined the Single Shape Fréchet Inception Distance (SSFID) for evaluating to what extent the generative model captures the patch statistics of the training shape. SSFID is defined as the FID between deep feature maps extracted from real and generated shapes.

SSIFD is commonly employed in models trained on a single 3D shape. For instance, \cite{wu_LearningGenerate3D_2022a} trains on a voxel pyramid of the input shape and operates over a hierarchy of tri-plane feature maps, enabling the model to produce diverse shape variations across different bounding boxes and aspect ratios, while preserving large-scale structural similarity to the input shape. Similarly, Sin3DM \cite{wu_Sin3DMLearningDiffusion_2023} learns the internal patch distribution from a single textured 3D shape to produce high-quality variations, and ShapeShifter \cite{maruani_ShapeShifter3DVariations_2025a} extends diffusion models with a multiscale pipeline operating on sparse voxel grids to achieve diverse yet structurally consistent 3D generations.

\subsection{Perceptual and Realism Metrics}

These metrics are designed to evaluate the perceptual or statistical realism of generated shapes, particularly in settings where no direct ground-truth reference is available for comparison.

\subsubsection{Fréchet PointNet Distance (FPD)}

Inspired by FID, tree-GAN \cite{shu_treeGAN3DPointCloud_2019} introduced the ``Fréchet Point Cloud Distance"--later referred to as Fréchet PointNet Distance--by replacing 2D image features with 3D point cloud features from a PointNet classifier. FPD thus serves as a 3D analogue of FID for evaluating the quality of generated point clouds. Subsequent works (e.g.~\cite{li2019pu, liu2021diffusion}) have used FPD, where a pretrained PointNet/PointNet++ embeds shapes and the Fréchet distance between real and generated means/covariances is computed. Lower FPD signifies closer alignment to the real shape distribution, analogous to FID.

FPD calculates the 2-Wasserstein distance between real and fake Gaussian measures in the feature spaces extracted by~\cite{qi_PointNetDeepLearning_2017} as follows:

\begin{equation}
\text{FPD}(\mathbb{P}, \mathbb{Q}) = \| \mathbf{m}_\mathbb{P} - \mathbf{m}_\mathbb{Q} \|_2^2 + \text{Tr}(\Sigma_\mathbb{P} + \Sigma_\mathbb{Q} - 2(\Sigma_\mathbb{P}\, \Sigma_\mathbb{Q})^{\frac{1}{2}}),
\end{equation}

where $\mathbf{m}_\mathbb{P}$ and $\Sigma_\mathbb{P}$ are the mean vector and covariance matrix of the points calculated from real point clouds $\{x\}$, respectively, and $\mathbf{m}_\mathbb{Q}$, $\Sigma_\mathbb{Q}$ are the mean vector and covariance matrix calculated from generated point clouds $\{x'\}$, respectively, where $x  \thicksim \mathbb{P}$ and $x' = G(z)  \thicksim \mathbb{Q}$. $Tr(a)$ is the sum of the elements along the main diagonal of a matrix A.

\subsubsection{CLIP-Score / FID-3D}

Proposed by~\cite{hessel_clipscore_2021} as a reference-free metric for image captioning using OpenAI's CLIP model. CLIP-Score computes the cosine similarity between an image and a text embedding, measuring how well a generated image matches a given description. For an image with visual CLIP embedding $v$ and a candidate caption with textual CLIP embedding $c$, $w=2.5$ is set up and computes CLIP-Score as: 

\begin{equation}
\text{CLIP-Score}(c,\, v) = w * max(cos(c,\, v),0),
\end{equation}

Building on this idea, GET3D~\cite{chen2022get3d} adapted the FID to the 3D domain, introducing FID-3D, which evaluates the realism of 3D shapes by rendering them into images and computing FID over the rendered outputs. In essence, FID-3D applies the Fréchet metric to distributions of rendered images (or learned 3D features), capturing both shape and texture quality in a perceptual space. 

CLIP-Score is used in text-to-3D generation to evaluate semantic alignment between a generated 3D shape (usually via its images) and the input text prompt. For example, Magic3D~\cite{lin2023magic3d} reports CLIP-based scores to compare fidelity to the prompt, and JointDreamer~\cite{chen2023jointdreamer} achieved high CLIP R-Precision (88.5\%) for their text-guided 3D outputs. FID-3D was employed in GET3D to quantitatively evaluate 3D textured shape generation, and later works (e.g.~\cite{gao2023text2tex}) use FID-3D to measure improvements in 3D shape and texture realism. Both metrics have become useful for 3D generative models that involve images or multi-modal output, connecting 3D generation to 2D image-based criteria.

\subsubsection{ULIP and Uni3D}

Building on the idea of CLIP-Score, ULIP \cite{xue_ULIPLearningUnified_2023} and Uni3D \cite{zhou_Uni3DExploringUnified_2023} have been proposed as cross-modal evaluation metrics to assess the alignment between generated 3D shapes and their corresponding textual or visual representations in a unified embedding space.

ULIP learns a joint representation of images, text, and 3D point clouds by pre-training on triplets of objects across the tree modalities. To mitigate the high demand of 3D data, ULIP leverages powerful pre-trained image and text encoders trained on massive image-text pairs, and subsequently aligns the 3D encoder to this pre-aligned image-text feature space using a small set of training triplets \cite{xue_ULIPLearningUnified_2023}. The construction of these triplets follows a three-stage pipeline: (i) First, CAD models from ShapeNet55 \cite{chang_ShapeNetInformationRich3D_2015} are uniformly sampled with $N_p$ points, and standard point cloud data augmentation are performed. Subsequently, a 3D encoder takes the augmented point cloud $P_i$ as input and outputs it 3D representation $\mathbf{h^p_i}$ via $\mathbf{h^p_i} = f_P(P_i)$, where $f_P(\cdot)$ represent the 3D backbone encoder (eg., PointNet++, PointBert and PointMLP). (ii) Since ShapeNet55 CAD models do not come with images, synthetic multi-view RGB images and depth maps are generated by placing visual cameras around each CAD model. From these, and image (or depth map) $I_i$ is randomly selected and encoded as $\mathbf{h^{{I}}_i} = f_I(I_i)$, where $f_I(\cdot)$ is the image encoder. (iii) Finally, text descriptions are obtained from the metadata associated with each CAD model to generate textual features.

Similarly, Uni3D introduces a large multi-modal constrastive learning framework that aligns images and language with 3D point clouds, thereby providing a unified feature space suitable for both representation learning and evaluation.

Recent methods such as Hunyuan3D 2.1 \cite{hunyuan3d_Hunyuan3D21Images_2025} and VolGen \cite{tang_VolGenVolumetricLatent_2025} adopt ULIP and Uni3D as evaluation metrics to measure cross-modal consistency between generated 3D assets and their conditioning inputs. In particular, Hunyuan3D 2.1 reports ULIP-I and ULIP-T scores, which quatify the similarity between the point cloud and text, as well as the similarity between the point cloud and image, respectively. Likewise, Uni3D-I and Uni3D-T scores are obtained by applying the same procedure using the Uni3D model

\section{Discussion and Future Directions}\label{sec:discussion}

After reviewing recent works on methods for 3D shape generation, in this section, we discuss the open challenges and future directions in this research field.

\subsection{Multi-Representation Learning for 3D Shape Generation}

As discussed in Section \ref{section_3d_shape_representations}, various 3D shape representations exist, including explicit, implicit, and, more recently, hybrid representations. Each representation, however, has inherent limitations that may affect either the preservation of geometric details, the flexibility of the shape structure, or the overall design of the generative model. 

Implicit functions, for instance, cannot explicitly represent or intuitively allow editing of output surfaces as easily as meshes~\cite{sun_RecentAdvancesImplicit_2024}. Conversely, mesh-based representation often requires a fixed topology, particularly in approaches relying on category-specific mesh models~\cite{zuffi_LionsTigersBears_2018, kolotouros_ConvolutionalMeshRegression_2019}. To address this bottleneck, methods such as ~\cite{groueix_AtlasNetPapierMacheApproach_2018, wang_Pixel2MeshGenerating3D_2018} extended mesh generation to a broader range of object categories by incorporating topological priors. While meshes are advantageous for preserving fine geometric details and enabling deformation, modifying their topology remains challenging.

Continuing along the line of 3D shape generation, works like~\cite{chen2022get3d} have revisited traditional ``mesh and texture" representation, achieving good performance by transforming implicit surfaces to explicit meshes using techniques such as Deep Marching Tetrahedra~\cite{shen_DeepMarchingTetrahedra_2021}.

In summary, leveraging the strengths while mitigating the weaknesses of different 3D representations remains a promising direction for future research.

\subsection{Challenges in achieving High-Quality 3D Representations}

Although existing works are capable of generating 3D shapes with desirable characteristics (e.g., symmetry), they still struggle to produce fine-grained geometric details. As discussed in Section \ref{section_3d_shape_representations}, generative networks based on voxel grids and implicit representations require massive computing resources to synthesize high-quality shapes with detailed geometry. In contrast, while point clouds are more memory-efficient, they demand a large number of points to faithfully represent a complete shape. Moreover, the lack of local topological connectivity and ambiguous global topology significantly affects the fidelity of the reconstructed 3D shapes.

Meshes, which naturally encode both geometry and topology, offer a more balanced representation. However, their inherent irregularity poses challenges in network design and often limits the precision of the generated details.

Beyond geometric fidelity, the appearance quality of generated shapes also remains an open challenge. Radiance fields and 2D-to-3D generation architectures show promise in jointly modeling appearance and geometry. Nevertheless, methods that provide fine-grained control over the appearance of generated shapes are still limited~\cite{sun_RecentAdvancesImplicit_2024}.

Based on these observations, further research is needed to improve the control over both the geometric and appearance quality of generated shapes, enabling generative models to capture finer details while maintaining coherent global structures.

\subsection{Network Design and Large-Scale Models for 3D Shape Generation}

Effective 3D shape generation requires backbone networks that can simultaneously encode complex 3D structures into latent embeddings and reconstruct detailed shapes for them~\cite{xu_SurveyDeepLearningbased_2023}. Developing specialized backbones has been crucial for advancing the quality and diversity of 3D shape generation. Historically, significant milestones in 3D generation have often been driven by innovation in network architectures, such as volumetric convolutional networks~\cite{zhirongwu_3DShapeNetsDeep_2015}, point-based networks like PointNet~\cite{qi_PointNetDeepLearning_2017}, and implicit function-based architectures like DeepSDF~\cite{park_DeepSDFLearningContinuous_2019}.

The choice of backbone architecture not only determines the underlying 3D representation but also heavily influences the generative modeling approach. For instance, voxel-based generators~\cite{wu_LearningProbabilisticLatent_2016} tend to be limited by the resolution constraints, while point cloud generators~\cite{achlioptas_learning_2018} must address issues of unordered structure and topology.

More recently, the emergence of large-scale models has begun to impact 3D shape generation directly. Researchers are moving beyond distilling knowledge from late 2D models to optimizing 3D content and are now focusing on training native large-scale 3D generative models~\cite{li_Advances3DGeneration_2024}. MeshGPT~\cite{siddiqui_MeshGPTGeneratingTriangle_2024}, for instance, adapts transformer-based architectures inspired by large language models to autoregressively generate 3D triangle meshes. However, despite the innovation, MeshGPT's generative capabilities remain constrained by the ability and diversity of 3D training datasets, primarily producing regular objects such as furniture.

To sum up, advances in backbone network design and the development of large-scale 3D generative models have significantly expanded the capabilities of 3D shape generation. Nevertheless, further research is needed to design a backbone that can efficiently handle various 3D representations, scale to large datasets, and generate more diverse and structurally complex 3D shapes.

\subsection{Evaluation Protocols for 3D Shape Generation}

Quantitatively evaluating the quality of generated 3D shapes remains a significant and relatively unexplored problem~\cite{li_Advances3DGeneration_2024}. Unlike the 2D domain, where evaluation metrics are well-established, 3D shape generation faces additional challenges due to the complexity of 3D data and the absence of standardized benchmarks.

Some evaluation metrics, such as fidelity-based measures (see Subsection \ref{sec:fidelity_metrics}), require ground-truth data. However, while they assess how closely a generated shape matches real samples, they often fail to capture broad aspects such as diversity, structural plausibility, or perceptual quality. This highlights the need for more comprehensive and multi-faceted evaluation protocols. 

In practice, some methods design their network architectures explicitly to optimize for specific evaluation metrics, while others propose new metrics tailored to their approaches. As a result, the field lacks a unified evaluation protocol, making fair comparisons across different methods challenging.

Furthermore, in contrast to text or 2D images--which are relatively easy to collect at scale--3D assets typically require manual creation by professional artists using specialized software, leading to high production costs. Consequently, the availability of large-scale, high-quality 3D datasets for training and evaluation remains limited.

In this context, there is a need to define standardized evaluation protocols grounded on widely adopted 3D shape datasets. Although ShapeNet, following the data preprocessing pipelines introduced in PointFlow, is commonly used for training and evaluation, many recent works rely on custom datasets or task-specific benchmarks. Addressing this inconsistency is essential for the field's progress, ensuring reproducibility, fair evaluation, and meaningful comparison across different 3D shape generation methods.

\subsection{Symmetry-aware 3D Generative Models}

\begin{figure*}[tbp]
\centering
\includegraphics[width=1\textwidth]{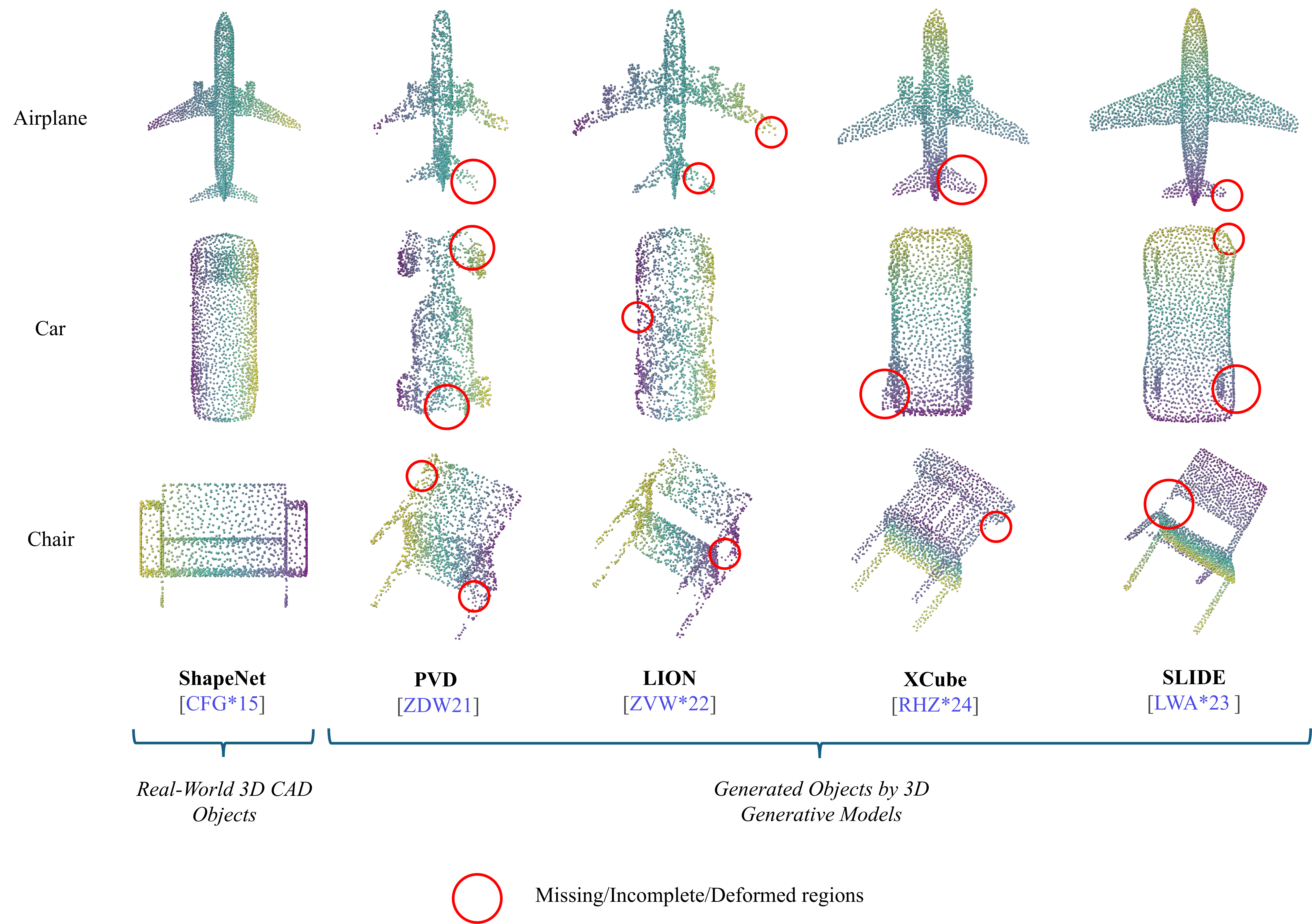}
\caption{(a) Reference shapes from the ShapeNet dataset: Airplane, Car, and Chair. (b) Unconditional shape generation with 2,048 points for each class-specific model trained on PointFlow's ShapeNet data with global normalization. Red circles highlight missing points and deformed regions.}
\label{fig:generated_shapes}
\end{figure*}

One of the main limitations of current generative models for 3D shapes is the lack of explicit guarantees regarding the preservation of geometric or structural properties present in the training data, such as symmetry, topological consistency, or surface continuity. Most approaches focus on learning an implicit distribution over shape space without enforcing constraints that preserve these properties~\cite{wang_Pixel2MeshGenerating3D_2018}.

This leads to a methodological gap: as can be seen in Figure \ref{fig:generated_shapes}, although many 3D objects present in widely used datasets exhibit intrinsic symmetry, generative models do not encode this property explicitly. In theory, models are expected to capture symmetry as a statistical pattern -- taking into account that many objects in the training datasets are symmetrical -- but there is no formal guarantee that this occurs in practice. Empirical evidence shows that generated shapes can be visually plausible yet structurally inconsistent or asymmetric.

These limitations call for explicit strategies to incorporate geometric priors during generation. Symmetry, in particular, is a well-understood property in shape analysis and a natural candidate for inductive bias. While recent works have explored symmetry-aware architectures~\cite{yang_SYM3DLearningSymmetric_2024, kelvinius_WyckoffDiffGenerativeDiffusion_2025, nagar_RobustExtrinsicSymmetry_2025}, the area remains underdeveloped. Understanding how current models handle symmetry is thus essential for designing methods that improve structural fidelity.

\subsection{Efficiency and Scalability in 3D Generative Models}

According to \cite{wu_Sin3DMLearningDiffusion_2023}, current 3D generative models rely on large-scale 3D datasets for training. However, collecting high-quality 3D data is much more expensive than images, and many artistically designed shapes have unique structures that are difficult to learn from limited collections. Moreover, state-of-the-art methods still struggle to reproduce the fine geometric details and sharp features required for digital shapes in geometry modeling \cite{maruani_ShapeShifter3DVariations_2025a}.

\tableTrainingTimesComp

As summarized in Table \ref{tab:trainingTimeComp}, conventional 3D generative models (top) require substantial computational resources and involve long training and inference times. To address these limitations, a new class of 3D generative models has emerged; in fact, \cite{wu_Sin3DMLearningDiffusion_2023} coined the concept of \textit{single instance generative models (SIGMs)}, which aim to learn the internal patch statistics from a single input instance and generate diverse new samples with similar local content.

Early attempts include DGTS \cite{hertz_DeepGeometricTexture_2020}, a GAN-based method that learns local geometric textures on mesh surfaces but not global structures, and SSG \cite{wu_LearningGenerate3D_2022a}, another GAN-based 3D shape generative model that only generates geometry from a single shape. Sin3DM \cite{wu_Sin3DMLearningDiffusion_2023} further extends these approaches by adding 3 extra dimensions for RGB color and training a diffusion model on a single 3D textured shape with locally similar patterns. Sin3DGen \cite{li_PatchBased3DNatural_2023} applies patch matching on the Plenoxels representation \cite{fridovich-keil_PlenoxelsRadianceFields_2022}.

More recently, ShapeShifter \cite{maruani_ShapeShifter3DVariations_2025a} employs a multiscale diffusion framework with limited receptive fields to learn the internal structures of a given shape, building upon \cite{kulikov_SinDDMSingleImage_2023}, an approach that has been used for training a generative model on a single image.

As also shown in Table \ref{tab:trainingTimeComp}, SIGMs (bottom) are considerably more efficient, requiring fewer computational resources and shorter training and inference times (decreasing from hundreds of GPU hours to a couple of hours or minutes on single GPUs).

We consider this a promising research direction, as it opens the possibility of high-quality 3D generation without the prohibitive costs associated with large-scale datasets and training times. By learning directly from a single example, these methods bypass the need for extensive curation pipelines, large storage infrastructures, and prolonged training and inference schedules that characterize dataset-driven approaches. Furthermore, SIGMs are particularly well-suited for domains where data is scarce, heterogeneous, or inherently unique, such as artist-designed assets, cultural heritage artifacts, or specialized industrial components. 

\section{Conclusions}

In this work, we present a comprehensive survey on 3D shape generation, encompassing three core components: 3D representations, generation methods, and evaluation protocols. We begin by introducing the major categories of 3D shape representation, which define the structure and properties of the generated shapes and play a central role in model design. Next, we summarize and categorize a wide range of generation methods based on the feedforward generation paradigm. Subsequently, we summarize the datasets used to train those generation methods and the metrics used to evaluate the generation quality of the generated shapes. Finally, we discuss current challenges and outline future research directions in the field of 3D shape generation. We hope this survey provides a comprehensive overview of 3D shape generation, serving as a valuable reference for both researchers and newcomers in this field.

\section*{Acknowledgements}

This work was supported by ANID Chile - Fondecyt Regular N° 1251263, the National Center for Artificial Intelligence CENIA FB210017, Basal ANID, and the Shape Vision Lab from the University of Chile.

\newpage
\bibliographystyle{eg-alpha-doi} 
\bibliography{references}       




\end{document}